\documentclass[10pt,twocolumn,letterpaper]{article}

\usepackage{cvpr}              

\usepackage{multirow} 
\usepackage{booktabs}
\usepackage[colorlinks]{hyperref}

\usepackage{bbding}
\usepackage{pifont}
\usepackage[table]{xcolor}
\usepackage{stfloats}
\definecolor{cvprblue}{rgb}{0.21,0.49,0.74}

\def\paperID{3136} 
\def\confName{CVPR}
\def\confYear{2024}

\title{PromptAD: Learning Prompts with only Normal Samples for \\ Few-Shot Anomaly Detection}

\author{Xiaofan Li$^1$, Zhizhong Zhang$^1$, Xin Tan$^{1,2,}$\thanks{Corresponding author.}, Chengwei Chen$^2$, Yanyun Qu$^3$, Yuan Xie$^{1,2}$, Lizhuang Ma$^1$\\
$^1$East China Normal University, Shanghai, China\\
$^2$Chongqing Institute of East China Normal University, Chongqing, China\\
$^3$The Navy Military Medical University, Shanghai, China\\
$^4$Xiamen University, Fujian, China\\
{\tt\small \{funzi\}@stu.ecnu.edu.cn, \{zzzhang, xtan, yxie, lzma\}@cs.ecnu.edu.cn} \\ {\tt\small \{timchen\}91@aliyun.com,  \{yyqu\}@xmu.edu.cn}
}

\begin{document}
\maketitle
\begin{abstract}
The vision-language model has brought great improvement to few-shot industrial anomaly detection, which usually needs to design of hundreds of prompts through prompt engineering. For automated scenarios, we first use conventional prompt learning with many-class paradigm as the baseline to automatically learn prompts but found that it can not work well in one-class anomaly detection. To address the above problem, this paper proposes a one-class prompt learning method for few-shot anomaly detection, termed PromptAD. First, we propose semantic concatenation which can transpose normal prompts into anomaly prompts by concatenating normal prompts with anomaly suffixes, thus constructing a large number of negative samples used to guide prompt learning in one-class setting. Furthermore, to mitigate the training challenge caused by the absence of anomaly images, we introduce the concept of explicit anomaly margin, which is used to explicitly control the margin between normal prompt features and anomaly prompt features through a hyper-parameter. For image-level/pixel-level anomaly detection, PromptAD achieves first place in 11/12 few-shot settings on MVTec and VisA. Code is available at \href{https://github.com/FuNz-0/PromptAD.git}{https://github.com/FuNz-0/PromptAD.git} 
\end{abstract}    
\section{Introduction}
\label{sec:intro}

Anomaly detection (AD) \cite{wu, MvTec, PatchCore} is a critical task in computer vision \cite{VS-Boost, En-compactness, hao1, hao2}, with widespread applications of defect detection in industry and medicine. This paper focuses on unsupervised industrial anomaly detection, which poses a challenge known as a one-class classification (OCC) \cite{OCC} setting. In this framework, only normal samples are available during training, but in the testing phase, the model is expected to identify anomalous samples. Since industrial anomaly detection typically customizes a model for various industrial production lines, the ability to rapidly train models with few samples holds significant promise for practical applications.

\begin{figure}[t]
  \centering
   \includegraphics[width=1\linewidth]{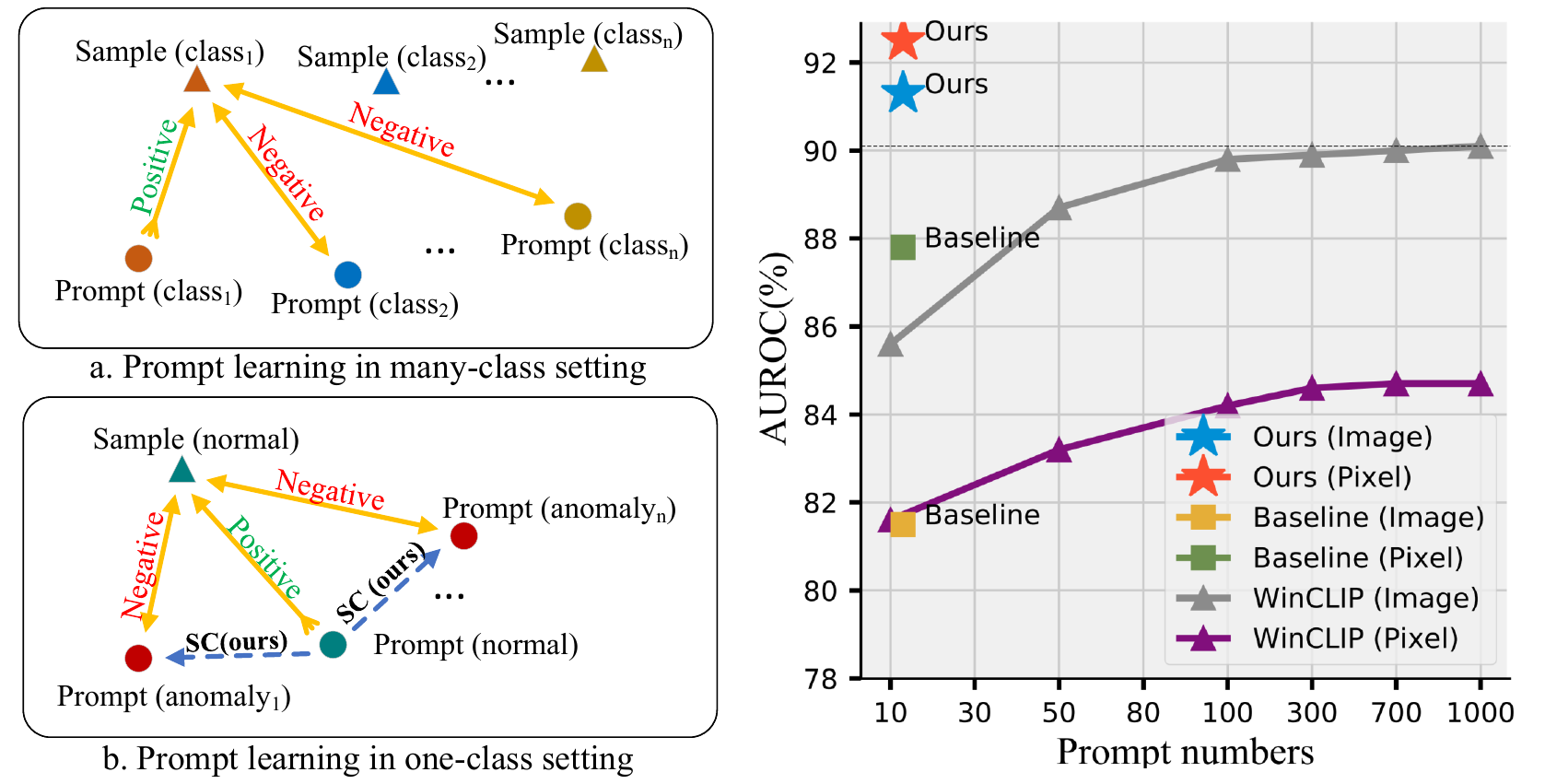}

   \caption{\textbf{Left:} Prompt learning under many-class and one-class settings. \textbf{Right:} The prompt-guided results of WinCLIP using different numbers of prompts, and the prompt-guided results of the baseline and our PromptAD under one-shot for prompt learning. All results are on the MVTec.}
   \label{fig:onecol}
   \vspace{-10pt}
\end{figure}

Due to the strong zero-shot ability of the foundation models \cite{GPT, LDM, BLIP}, WinCLIP \cite{WinCLIP} was proposed as the first work utilizing the vision-language foundation model (\ie, CLIP \cite{CLIP}) to enhance the model's anomaly detection performance in few-shot settings. To better leverage prompt guidance, WinCLIP introduces a prompt engineer strategy called ``Prompt Ensemble'' which combines a sufficient number of manually-designed prompts. For example, some manual prompts (\eg, \verb'a cropped photo of a []', \verb'a blurry photo of the []', \etc) are combined together as the normal prompts. As shown in Figure \ref {sec:intro} (right), with the number of prompts increasing, WinCLIP's performance improves, reaching a saturation point at around 1000 prompts. Other methods like SAA+ \cite{SAA} and AnoVL \cite{AnoVL} also employ prompt engineering to enhance model performance, which has become a rite of prompt-guided anomaly detection. Prompt engineering involves human intervention and requires careful design, which does not meet the automation requirements of industrial scenarios. 

As illustrated in Fig.\ref{sec:intro} (left a.), prompt learning \cite{COOP} aims to automatically learn prompts through contrastive learning \cite{MOCO, SimCLR} for guiding image classification. The idea of prompt learning for anomaly detection is intriguing. However, as shown in Figure \ref {sec:intro} (right), due to the one-class setting of anomaly detection, using the above prompt learning paradigm \cite{COOP} as the baseline does not work well and is inferior to WinCLIP \cite{WinCLIP} with manual prompts on the image-level result. The main challenges are as follows: 1) prompt learning relies on contrastive learning, how to design prompts to complete the contrastive learning in the one-class setting? 2) With the absence of anomaly samples, how to control the marginal distance between normal prompts and anomaly prompts? 

In this paper, we propose the one-class prompt learning with only normal samples for AD termed \textbf{PromptAD}. To solve the first challenge above, we propose semantic concatenation (SC). Intuitively, concatenating a prompt with antisense texts can transpose its semantics. According to this idea, as illustrated in Figure \ref{sec:intro} (left b.), SC first designs a learnable normal prompt such as $[\textbf{P}_1][\textbf{P}_2]\dots[\textbf{P}_{E_N}][obj.]$ for normal samples, and then manually concatenate various texts related to anomalies with the normal prompt such as $[\textbf{P}_1][\textbf{P}_2]\dots[\textbf{P}_{E_N}][obj.][with][flaw]$ which is converted into an anomaly prompt and can be used as a negative prompt of normal sample during prompt learning. Due to the manually annotated anomalous texts are very limited. To expand the richness of anomaly information, SC also designs learnable anomaly prompts by concatenating a suffix of learnable tokens with a normal prompt, for instance $[\textbf{P}_1][\textbf{P}_2]\dots[\textbf{P}_{E_N}][obj.][\textbf{A}_{1}][\textbf{A}_{2}]\dots[\textbf{A}_{E_A}]$, where $[\textbf{A}_i]$ is learnable token. The distribution of learnable anomaly prompts and manual anomaly prompts are aligned to ensure that the learnable anomaly prompts learn more correct anomaly information. 
 
Furthermore, in anomaly detection, anomaly samples are unavailable, making it impossible to explicitly control the margin between normal and anomaly prompt features through contrastive loss. To address the second challenge, we propose the concept of Explicit Anomaly Margin (EAM), where a hyper-parameter is introduced to ensure that the distance between normal features and normal prompt features is smaller than the distance between normal features and anomaly prompt features. Thus ensuring a sufficient margin between normal prompts and anomaly prompts. Figure \ref {sec:intro} (right) illustrates our great advantages, it can be seen that (compared with the WinCLIP \cite{WinCLIP} and Baseline \cite{COOP}) PromptAD achieves 91.3\%($\uparrow$1.2\% and $\uparrow$9.8\%)/92.5\%($\uparrow$7.7\% and $\uparrow$3.7\%) image-level/pixel-level anomaly detection results with only 10$\sim$20 ($\downarrow$ $\sim$980 and $\downarrow$ 0) prompts.

To summarize, the main contributions of this paper are:

\begin{itemize}
    \item We explore the feasibility of prompt learning in one-class anomaly detection, and propose a one-class prompt learning method termed \textbf{PromptAD}, which thoroughly beats conventional many-class prompt learning.

    \item Semantic concatenation (SC) is proposed, which can transpose the semantics of normal prompts by concatenating anomaly suffixes, so as to construct enough negative prompts for normal samples.

    \item Explicit anomaly margin (EAM) is proposed, which can explicitly control the distance between normal prompt features and anomaly prompt features through a hyper-parameter.
	
    \item For image-level/pixel-level anomaly detection, PromptAD achieves first place in 11/12 few-shot settings on MVTec \cite{MvTec} and VisA \cite{Visa}.
\end{itemize}

\begin{figure*}[t]
  \centering
   \includegraphics[width=1.0\linewidth]{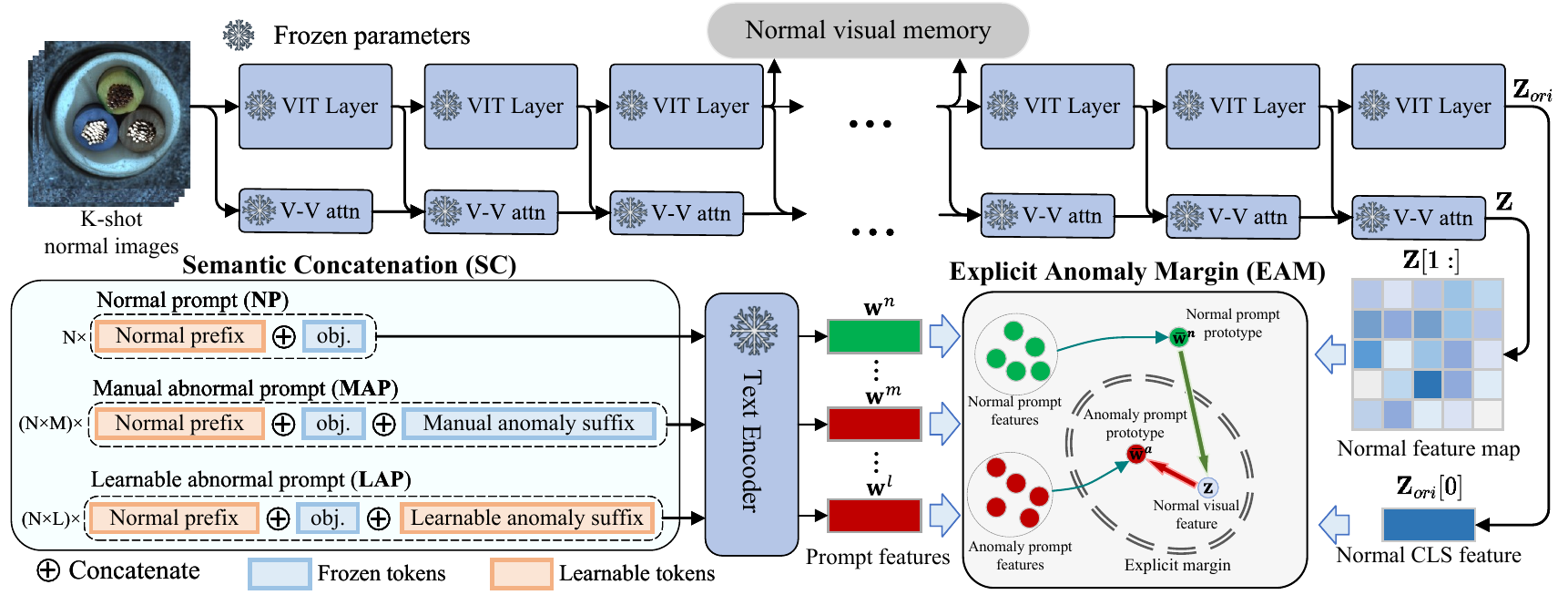}
    
   \caption{Illustration of PromptAD, which includes two novel modules: SC and EAM. The visual encoder has been transformed with v-v attention. The original branch is used to extract CLS feature, while the v-v attention branch is used to extract the feature map.}
   \label{method}
   \vspace{-10pt}
\end{figure*}
\section{Related Work}
\noindent \textbf{Vision-Language Model.} Leveraging contrastive learning \cite{MOCO, SimCLR} and vision transformer \cite{Visa}, some vision-language models (VLM) \cite{CLIP, BLIP, VLMo, ALIGN} have recently achieved great success. CLIP is one of the most commonly used VLMs, which is trained on web-scale image-text and shows strong zero-shot classification ability. The code of CLIP for LAION-400M \cite{Laion-400m} and LAION-5B \cite{Laion-5b} scale pre-training is open-scoured by OpenCLIP \cite{OpenCLIP}. With the pre-trained CLIP and prompt engineer, huge leaps were made for some downstream tasks \cite{CLIP-Reid, ZegCLIP, CRIS, SP-CLIP}. Influenced by the success of prompt learning \cite{PL-NLP1, PL-NLP2} in Natural Language Processing (NLP), there has been a surge of prompt learning methods \cite{COOP, COCOOP, CLIP-Adapter, SADA} in recent times for few-shot image classification tasks. These methods aim to automatically learn better prompts through contrastive learning \cite{MOCO, SimCLR} for guiding image classification based on CLIP.

\vspace{0.5\baselineskip}
\noindent \textbf{Anomaly Detection.} Most of the AD methods mainly focus on three paradigms: feature embedding paradigm, knowledge distillation paradigm, and reconstruction-based paradigm. The feature embedding paradigm \cite{InReaCh, PaDiM, PNI, PatchCore, SPADE, PatchSVDD, SimpleNet, DifferNet} extracts the patch features of the image through the neural network and then performs anomaly detection,. The knowledge distillation paradigm \cite{MKD, EfficientAD, DeSTSeg, FPN, MemKD, FPN} lets the student network only learn the knowledge of the normal samples of the teacher network, and complete anomaly detection through the difference between the teacher and the student. The reconstruction paradigm \cite{THFR, Draem, DiffAD} hopes that the model can reconstruct the anomaly image into a normal image, and realize anomaly detection by the difference between the reconstructed image and the anomaly image.

\vspace{0.5\baselineskip}
\noindent \textbf{Few-Shot Anomaly Detection.} TDG \cite{TDG} and RegAD \cite{RegAD} are the first to explore few-shot anomaly detection methods, and PatchCore \cite{PatchCore} and DifferNet \cite{DifferNet} also demonstrated the performance in few-shot settings. WinCLIP \cite{WinCLIP} and RWDA \cite{RWDA} introduce the CLIP model to anomaly detection and greatly improve the performance in the few-shot setting. The latest FastRecon \cite{FastRecon} reconstructs anomaly features by regression with distribution regularization and achieves excellent performance.

\section{Preliminaries}

\subsection{CLIP and Prompt Learning}
\textbf{Contrastive Language Image Pre-training} termed CLIP \cite{CLIP} is a large-scale vision-language model which is famous for its zero-shot classification ability. Specifically, giving an unknown image $\textbf{i}$, and $K$ text-prompts $\{\textbf{s}_1, \textbf{s}_2, ..., \textbf{s}_K \}$, CLIP can predict the distribution of $\textbf{i}$ belonging to these $K$ text-prompts:
\begin{equation}
    \begin{aligned}
        p(\textbf{y}|\textbf{i}) = \frac{\textup{exp}<f(\textbf{i}), g(\textbf{s}_y)/\tau>}{\sum_{i=1}^K \textup{exp}<f(\textbf{i}), g(\textbf{s}_i)/\tau>},
    \end{aligned}
\label{eq:zsl}
\end{equation} 
where $f(\cdot)$ and $g(\cdot)$ are visual and text encoder respectively. $<\cdot, \cdot>$ represents cosine similarity, $\tau$ is the temperature hyper-parameter. The initial text prompt used for CLIP zero-shot classification is still simple, such as \verb'a photo of [class]', \etc, slightly better than directly using the name of the class as the prompt. 


\vspace{0.5\baselineskip}
\noindent\textbf{Prompt Learning.} Inspired by the success of prompt learning in natural language processing (NLP) \cite{PL-NLP1, PL-NLP2}, CoOp \cite{COOP} introduces this paradigm into few-shot classification, aiming to automatically learn efficient prompts for CLIP. Specifically, the prompt used in CoOp is not the frozen text description, but a set of trainable parameters:
\begin{equation}
    \begin{aligned}
        \textbf{s}_k = [\textbf{P}_1][\textbf{P}_2] \dots [\textbf{P}_{E_P}][class_k],
    \end{aligned}
\label{eq:prompt}
\end{equation} 
where $[\textbf{P}_1][\textbf{P}_2] \dots [\textbf{P}_{E_P}]$ are trainable tokens and $[class_k]$ is $k$-th class name which is not trainable. Prompt learning aims to automatically train effective prompts to improve CLIP performance on downstream classification tasks.
\subsection{CLIP Surgery}
As a classification model, CLIP is far less adaptive in prompt-guided image localization tasks without fine-tuning. To find out why CLIP fails to image localization tasks, some CLIP explainable works \cite{MaskCLIP, CLIPSurgery} analyze the mechanism that how CLIP extracts visual features. These works observed that the global feature extraction of Q-K self-attention \cite{Transformer} affects the localization ability of CLIP, which is as follows:
\begin{equation}
    \begin{aligned}
        Attn(\textbf{Q},\textbf{K},\textbf{V})=softmax(\textbf{Q} \cdot \textbf{K}^\text{T} \cdot scale) \cdot \textbf{V}.
    \end{aligned}
\label{eq:qkv}
\end{equation} 
To this end, CLIP-Surgery \cite {CLIPSurgery} proposes a V-V attention mechanism to enhance the model's attention to local features without destroying the original structure. As shown in Figure \ref {method}, the feature extraction process is described as follows: 
\begin{equation}
    \begin{aligned}
        \textbf{Z}^{l-1}_{ori} = [\textbf{t}_{cls};\textbf{t}_1;\textbf{t}_2,...;\textbf{t}_T],
    \end{aligned}
\end{equation} 
\begin{equation}
    \begin{aligned}
        \textbf{Z}^{l-1} = [\textbf{t}_{cls}';\textbf{t}_1';\textbf{t}_2',...;\textbf{t}_T'],
    \end{aligned}
\end{equation} 
\begin{equation}
    \begin{aligned}
        [\textbf{Q}^l,\textbf{K}^l,\textbf{V}^l]=QKV\_Proj.^{l}(\textbf{Z}^{l-1}_{ori}),
    \end{aligned}
\end{equation} 
\begin{equation}
    \begin{aligned}
        \textbf{Z}^{l} = Proj.^{l}(Attn(\textbf{V}^l,\textbf{V}^l,\textbf{V}^l))+\textbf{Z}^{l-1},
    \end{aligned}
\end{equation} 
where $\textbf{Z}_{ori}^{l-1}$ denotes the ${(l-1)}$-th layer output of the original CLIP visual encoder and $\textbf{Z}^{(l-1)}$ denotes the local-aware output of layer $l-1$, $QKV\_Proj.^l$ and $Proj^l$ denote the QKV projection and output projection whose parameters are initialized by the visual encoder parameters of the original CLIP. The final original outputs and local-aware outputs are $\textbf{Z}_{ori}$ and $\textbf{Z}$, the CLS feature $\textbf{Z}_{ori}[0] \in \mathbb{R}^{d}$ is used for image-level anomaly detection and the local feature map $\textbf{Z}[1:] \in \mathbb{R}^{T\times d}$ is used for pixel-level anomaly detection. In this paper, we use modified CLIP as the backbone and term it VV-CLIP.

\section{Methodology}

\subsection{Overview}
An overview of our proposed PromptAD is illustrated in Figure \ref {method}. PromptAD is built on VV-CLIP whose visual encoder is used to extract global and local features. The proposed semantic concatenation (SC) is used to design prompts. Specifically, $N$ learnable normal prefixes and the objective name are concatenated to get normal prompts (NPs), then $N$ normal prompts are concatenated with $M$ manual anomaly suffixes and $L$ learnable anomaly suffixes respectively to obtain $N\times M$ manual anomaly prompts (MAPs) and $N\times L$ learnable anomaly prompts (LAPs). The visual features and prompt features are used to complete prompt learning by contrastive loss and the proposed explicit anomaly margin (EMA) loss. EMA can control the explicit margin between the normal prompt features and anomaly prompt features through a hyper-parameter. Finally, the prompts obtained by prompt learning are used for prompt-guided anomaly detection (PAD).

In addition to PAD, referring to WinCLIP+ \cite{WinCLIP}, we also introduce vision-guided anomaly detection (VAD). Specifically, as shown in Figure \ref {method}, during training, the $i$-th layer features (without CLS feature) output by the visual encoder are stored as normal visual memory which is denoted as $\textbf{R}$. In the testing phase, the $i$th layer feature map $\textbf{F} \in \mathbb{R}^{h \times w \times d}$ of a query image is compared with $\textbf{R}$ to obtain the anomaly score map $\textbf{M}\in [1,0]^{h\times w}$:
\begin{equation}
	\begin{aligned}
        \textbf{M}_{ij}=\min_{\textbf{r}\in \textbf{R}}\frac{1}{2}(1-<\textbf{F}_{ij}, \textbf{r}>).
	\end{aligned}
	\label{tip}
\end{equation}
In practice, we use the intermediate features of two layers as memory to get two score maps for each query image and then average the two score maps to get the final vision-guided score map $\textbf{M}_{v}$. 

\subsection{Semantic Concatenation}

Only normal samples are obtainable during anomaly detection training, which leads to no negative samples for guiding prompt learning and thus impairs its effect. We found that the semantics of prompts can be changed by concatenating. For example, \verb'a photo of cable' has normal semantics, and after concatenating it with a suffix, \verb'a photo of cable with flaw' is converted into anomaly semantics. In this way, we propose semantic concatenation (SC) which can transpose normal prompts to anomaly prompts by concatenating normal prompts with anomaly suffixes, so as to construct sufficient contrast prompts based on learnable normal prompts. Specifically, following the format of CoOp \cite{COOP}, the learnable normal prompt (NP) is designed as follows:
\begin{equation}
    \begin{aligned}
        \textbf{s}^{n} = [\textbf{P}_1][\textbf{P}_2] \dots [\textbf{P}_{E_N}][obj.],
    \end{aligned}
\label{eq:NP}
\end{equation} 
where $E_N$ denotes the length of the learnable normal prefix and $[obj.]$ represents the name of the object being detected. The learnable normal prompt can be transposed to an anomaly prompt after concatenating with the anomaly suffixes. In particular, we generated anomaly suffixes from the anomaly labels of the datasets \cite{MvTec, Visa}, such as \verb'[] with color stain', \verb'[] with crack', \etc, and then concatenate these texts with the NP to obtain the manual anomaly prompt (MAP):
\begin{equation}
    \centering
    \begin{aligned}
        \textbf{s}^{m} = [\textbf{P}_1][\textbf{P}_2] \dots [\textbf{P}_{E_N}][obj.][with][color][stain],\\
    \end{aligned}
\label{eq:MAP}
\end{equation} 
where the prefix is a trainable NP and the suffix is a manual anomaly text. In addition, we combine NP with a learnable token suffix to design the learnable anomaly prompt (LAP):
\begin{equation}
    \begin{aligned}
        \textbf{s}^{l} = [\textbf{P}_1][\textbf{P}_2] \dots [\textbf{P}_{E_N}][obj.][\textbf{A}_1]\dots[\textbf{A}_{E_A}],
    \end{aligned}
\label{eq:prompt}
\end{equation} 
where $E_A$ denotes the length of learnable anomaly suffix. It should be noted that the parameters of prompts concatenated by the same normal prefix or anomaly suffix are shared. During training, NPs move close to normal visual features, while MAPs and LAPs move away from normal visual features. The training loss for prompt learning is consistent with the CLIP training loss as follows:
\begin{equation}
    \begin{aligned}
    \mathcal{L}_{clip}=\mathbb{E}_{\textbf{z}}\left[-log\frac{\textup{exp}(\textless\textbf{z}, \bar{\textbf{w}}^n/\tau\textgreater)}{\textup{exp}<\textbf{z}, \bar{\textbf{w}}^n/\tau>+\sum\limits_{\textbf{w} \in \mathcal{W}} \textup{exp}<\textbf{z}, \textbf{w}/\tau>}\right],
    \end{aligned}
\label{eq:clip}
\end{equation} 
where $\textbf{z}$ denotes normal visual feature, $\bar{\textbf{w}}^n=\frac{\sum_{i=1}^{N}g(\textbf{s}^n_i)}{N}$ is the prototype of normal prompt features, $\mathcal{W}=\{g(\textbf{s})|\textbf{s}\in \textup{MAPs} \cup \textup{LAPs}\}$ is a set containing all anomaly prompt features. Since more negative samples can produce a better contrastive learning effect \cite{MOCO}, each anomaly prompt feature is compared with the visual feature.

\vspace{0.5\baselineskip}
\noindent \textbf{Remark.} In the one-class anomaly detection, conventional prompt learning can only design learnable normal prompts, which is not conducive to the effect of contrastive loss. The proposed semantic concatenation can transform the semantics of normal prompts into anomaly semantics with shared parameters, which can make normal samples contrast with the semantic transposes (anomaly prompts). 

\subsection{Explicit Anomaly Margin}
Due to the lack of anomaly visual samples in the training, the MAPs and LAPs can only take normal visual features as negative samples for contrast and lack an explicit margin between the normal and anomaly prompts. Therefore, we propose the explicit anomaly margin (EAM) for AD prompt learning, which can control the margin between normal prompt features and anomaly prompt features. EAM is actually a regularization loss implemented via a margin hyper-parameter, which is defined as:
\begin{equation}
	\begin{aligned}
        \mathcal{L}_{ema}=\mathbb{E}_\textbf{z}\left[
        \max \left(0,d(\frac{\textbf{z}}{\|\textbf{z}\|_2}, \frac{\bar{\textbf{w}}^n}{\|\bar{\textbf{w}}^n\|_2})-d(\frac{\textbf{z}}{\|\textbf{z}\|_2}, \frac{\bar{\textbf{w}}^a}{\|\bar{\textbf{w}}^a\|_2} )\right)
        \right],
	\end{aligned}
	\label{tip}
\end{equation}
where $d(\cdot, \cdot)$ represents euclidean distance, and $\bar{\textbf{{w}}}^a$ is the prototype of all anomaly prompt features:
\begin{equation}
	\begin{aligned}
        \bar{\textbf{{w}}}^a=\frac{\sum_{i=1}^{N\times M}g(\textbf{s}^m_i)+\sum_{i=1}^{N\times L}g(\textbf{s}^l_i)}{N\times M + N \times L}.
	\end{aligned}
	\label{tip}
\end{equation}
In CLIP, the final features are all projected onto the unit hyper-sphere, thus the features in $\mathcal{L}_{ema}$ are also normalized, and the margin is fixed to zero. Compared to contrastive loss ($\mathcal{L}_{clip}$), EMA loss guarantees a larger distance between normal samples and the anomaly prototype than between normal samples and the normal prototype, resulting in an explicit discrimination between normal and anomaly prototypes.

In addition, since MAPs contain sufficient anomaly information while LAPs are initialized without any semantic guidance, aligning them helps LAPs to mimic the distribution of MAPs. Specifically, we align the means of the two distributions using the squared $l_2$ norm:
\begin{equation}
	\begin{aligned}
        \mathcal{L}_{align}=\lambda \cdot \left\|\frac{\bar{\textbf{w}}^{m}}{\|\bar{\textbf{w}}^{m}\|_2}-\frac{\bar{\textbf{w}}^{l}}{\|\bar{\textbf{w}}^{l}\|_2}\right\|_2^2,
	\end{aligned}
	\label{tip}
\end{equation}
where $\bar{\textbf{w}}^m$ and $\bar{\textbf{w}}^l$ are the feature means of MAPs and LAPs, respectively, and $\lambda$ is a hyper-parameter controlling the alignment degree of MAPs and LAPs.
\subsection{Anomaly Detection}
In the testing phase, $\bar{\textbf{w}}^n$ is used as the normal prototype and $\bar{\textbf{w}}^a$ is used as the anomaly prototype to complete prompt-guided anomaly detection. The image-level score $\textbf{S}_t \in [0,1]$ and pixel-level score map $\textbf{M}_t \in [0,1]^{h\times w}$ are calculated through:
\begin{equation}
    \begin{aligned}
        score = \frac{\textup{exp}<\textbf{z}_t, \bar{\textbf{w}}^n/\tau>}{\textup{exp}<\textbf{z}_t, \bar{\textbf{w}}^n/\tau> + \textup{exp}<\textbf{z}_t, \bar{\textbf{w}}^a/\tau> },
    \end{aligned}
\label{eq:zsl}
\end{equation} 
where $\textbf{z}_t$ is a global/local image feature for image-level/pixel-level anomaly detection.

Finally, vision-guided $\textbf{M}_{v}$ and prompt-guided $\textbf{M}_t$ are fused to obtain the pixel-level anomaly score map, and the maximum value of $\textbf{M}_{v}$ and $\textbf{S}_t$ are fused to obtain the image-level anomaly score:
\begin{equation}
	\begin{aligned}
        \textbf{M}_{pix}=1.0 / (1.0 / \textbf{M}_{v} + 1.0 / \textbf{M}_{t}),
	\end{aligned}
	\label{tip}
\end{equation}
\begin{equation}
	\begin{aligned}
        \textbf{S}_{img}=1.0 / (1.0 / \max_{ij}\textbf{M}_{v} + 1.0 / \textbf{S}_{t}),
	\end{aligned}
	\label{tip}
\end{equation}
where the fusion method we use is harmonic mean, which is more sensitive to smaller values \cite{WinCLIP}.

\section{Experiments}
\label{sec:experiments}
We complete the comparison experiments between PromptAD and the latest methods under 1, 2, and 4-shot settings, which include both image-level and pixel-level results. In addition, we also compare the many-shot and full-shot methods to show the powerful few-shot performance of PromptAD. Finally, we conduct ablation experiments to verify the improvement of prompt learning by the proposed SC and EAM, and show the impact of different CLIP transformation methods \cite{CLIPSurgery, MaskCLIP} and hyper-parameters.
\vspace{0.5\baselineskip}

\noindent \textbf{Dataset.} In this paper, the benchmarks we use are MVTec \cite{MvTec} and VisA \cite{Visa}. Both benchmarks contain multiple subsets with only one object per subset. MVTec contains 15 objects with $700^2-900^2$ pixels per image, and VisA contains 12 objects with roughly $1.5\textup{K}\times1\textup{K}$ pixels per image. Anomaly detection is a one-class task, so the training set contains only normal samples, while the test set contains normal samples and anomaly samples with image-level and pixel-level annotations. In addition, the anomaly category present for each object is also annotated.
\vspace{0.5\baselineskip}

\noindent \textbf{Evaluation metrics.} We follow the literature \cite{MvTec} in reporting the Area Under the Receiver Operation Characteristic (AUROC) for both image-level and pixel-level anomaly detection.
\vspace{0.5\baselineskip}

\noindent \textbf{Implementation details.} We used the OpenCLIP \cite{OpenCLIP} implementation of CLIP and its pre-trained parameters, in addition to the default values of the hyper-parameter $\tau$. Referring to  WinCLIP \cite{WinCLIP}, we used LAION-400M \cite{Laion-400m} based CLIP with ViT-B/16+.

\begin{table*}[]
\centering
\scalebox{0.75}{
\setlength{\tabcolsep}{7mm}
\begin{tabular}{llcccccc}
\toprule
\multicolumn{1}{l}{\multirow{2}{*}{Method}} & \multicolumn{1}{l}{\multirow{2}{*}{Public}} & \multicolumn{3}{c}{MVTec}        & \multicolumn{3}{c}{VisA}       \\
\cmidrule(r){3-5} \cmidrule(r){6-8}
& & 1-shot   & 2-shot    & 4-shot    & 1-shot   & 2-shot   & 4-shot   \\ \midrule
SPADE \cite{SPADE}                      & arXiv'2020                 & 81.0\footnotesize{$\pm$2.0} & 82.9\footnotesize{$\pm$2.6}  & 84.8\footnotesize{$\pm$2.5}  & 79.5\footnotesize{$\pm$4.0} & 80.7\footnotesize{$\pm$5.0} & 81.7\footnotesize{$\pm$3.4} \\
PaDiM \cite{PaDiM}                      & ICPR'2020                  & 76.6\footnotesize{$\pm$3.1} & 78.9\footnotesize{$\pm$3.1}  & 80.4\footnotesize{$\pm$2.4}  & 62.8\footnotesize{$\pm$5.4} & 67.4\footnotesize{$\pm$5.1} & 72.8\footnotesize{$\pm$2.9} \\
PatchCore \cite{PatchCore}                & CVPR'2022                  & 83.4\footnotesize{$\pm$3.0} & 86.3\footnotesize{$\pm$3.3}  & 88.8\footnotesize{$\pm$2.6}  & 79.9\footnotesize{$\pm$2.9} & 81.6\footnotesize{$\pm$4.0} & 85.3\footnotesize{$\pm$2.1} \\
WinCLIP+\dag \cite{WinCLIP}                  & CVPR'2023                  & 93.1\footnotesize{$\pm$2.0} & \underline{94.4\footnotesize{$\pm$1.3}}  & \underline{95.2\footnotesize{$\pm$1.3}}  & \underline{83.8\footnotesize{$\pm$4.0}} & 84.6\footnotesize{$\pm$2.4} & \underline{87.3\footnotesize{$\pm$1.8}} \\
RWDA\dag \cite{RWDA}                & BMVC'2023                  & \underline{93.3\footnotesize{$\pm$0.5}} & 94.0\footnotesize{$\pm$0.7}  & 94.5\footnotesize{$\pm$0.7}  & 83.4\footnotesize{$\pm$1.7} & \underline{85.6\footnotesize{$\pm$1.4}} & 86.6\footnotesize{$\pm$0.9} \\
FastRcon \cite{FastRecon}                  & ICCV'2023                  & -        & 91.0 & 94.2 & -        & -        & -        \\ \midrule
\multicolumn{1}{l}{\textbf{PromptAD\dag}}    & \multicolumn{1}{c}{-}      & \textbf{94.6\footnotesize{$\pm$1.7}} & \textbf{95.7\footnotesize{$\pm$1.5}} & \textbf{96.6\footnotesize{$\pm$0.9}} & \textbf{86.9\footnotesize{$\pm$2.3}} & \textbf{88.3\footnotesize{$\pm$2.0}} & \textbf{89.1\footnotesize{$\pm$1.7}} \\ \bottomrule
\end{tabular}
}
\caption{Comparison of image-level anomaly detection in AUROC on MVTec and VisA benchmarks. The best and second-best results are respectively marked in bold and underlined. $\dag$ indicates CLIP-based methods.}
\label{image-level}
\vspace{-10pt}
\end{table*}

\subsection{Image-level Comparison Results}

The Image-level comparative experimental results of PromptAD and current methods are recorded in Table \ref {image-level}, where SPADE \cite{SPADE}, PaDiM \cite{PaDiM}, and PatchCore \cite{PatchCore} are the reformulations of traditional full-shot methods in the few-shot settings. It can be seen that the Image-level AD performance of these methods is very limited. Both WinCLIP+ \cite{WinCLIP} and RWDA \cite{RWDA} introduce CLIP \cite{CLIP}, which greatly improves the performance of Image-level AD under few-shot settings. Compared with the above methods, PromptAD achieves significant improvement in three settings of the two benchmarks. Compared with WinCLIP+ and RWDA, PromptAD achieves 1.3\%, 1.3\%, and 1.4\% (2.9\%, 2.7\%, 1.8\%) improvement under the 1, 2, and 4-shot Settings of MVTec (and VisA), respectively. In addition, PromptAD uses a smaller number of prompts than WinCLIP+ and RWDA.

\begin{table}[]
\centering
\scalebox{0.75}{
\setlength{\tabcolsep}{3.5mm}
\begin{tabular}{llccc}
\toprule
\multicolumn{1}{c}{Method} & Public    & Setting     & image                 & pixel                 \\ \midrule
\textbf{PromptAD} & - & 1-shot & 94.6 & 95.9 \\
\textbf{PromptAD} & - & 4-shot    & 96.6 & 96.5 \\ \midrule
DiffNet \cite{DifferNet} & WACV'2021 & 16-shot   & {\color[HTML]{FE0000} 87.3} & - \\
TDG \cite{TDG} & ICCV'2021 & 10-shot   & {\color[HTML]{FE0000} 78.0} & - \\
RegAD \cite{RegAD} & ECCV2022 & 8-shot    & {\color[HTML]{FE0000} 91.2} & 96.7 \\
FastRecon \cite{FastRecon} & ICCV'2023 & 8-shot    & {\color[HTML]{3166FF} 95.2} & 97.3 \\ \midrule
MKD \cite{MKD} & CVPR'2021 & full-shot & {\color[HTML]{FE0000} 87.8} & {\color[HTML]{FE0000} 90.7} \\
P-SVDD \cite{PatchSVDD} & ACCV'2021 & full-shot & {\color[HTML]{3166FF} 95.2} & {\color[HTML]{3166FF} 96.0} \\
PatchCore \cite{PatchCore} & CVPR'2022 & full-shot & 99.1 & 98.1 \\
SimpleNet \cite{SimpleNet} & CVPR'2023 & full-shot & 99.6 & 98.1 \\ \bottomrule
\end{tabular}
}
\caption{Comparison with exiting many-shot methods in AUROC (image and pixel level) on MVTec. Results below our 1-shot are marked in \textcolor{red}{red}, and those below our 4-shot are marked in \textcolor{blue}{blue}.}
\label{many-shot}
\vspace{-10pt}
\end{table}

\subsection{Pixel-level Comparison Results}
The pixel-level comparative experimental results are recorded in Table \ref {pixel-level}. It can be seen that the CLIP-based method (WinCLIP+ \cite{WinCLIP}) and other methods perform comparably on pixel-level AD, and the improvement brought by the introduction of CLIP \cite{CLIP} is not as obvious as that on image-level AD. PromptAD achieves the best place on MVTec/VisA in the 1-shot and 2-shot settings, which are 0.7\%/0.3\% and 0.2\%/0.3\% higher than WinCLIP+, respectively. In the 4-shot setting, while PromptAD ranks first on VisA, it takes second place on MVTec, narrowly outperformed by FastRecon \cite{FastRecon} with a 0.5\% margin. 

\begin{figure*}[t]
  \centering
  \begin{subfigure}{0.49\linewidth}
    \includegraphics[width=1\linewidth]{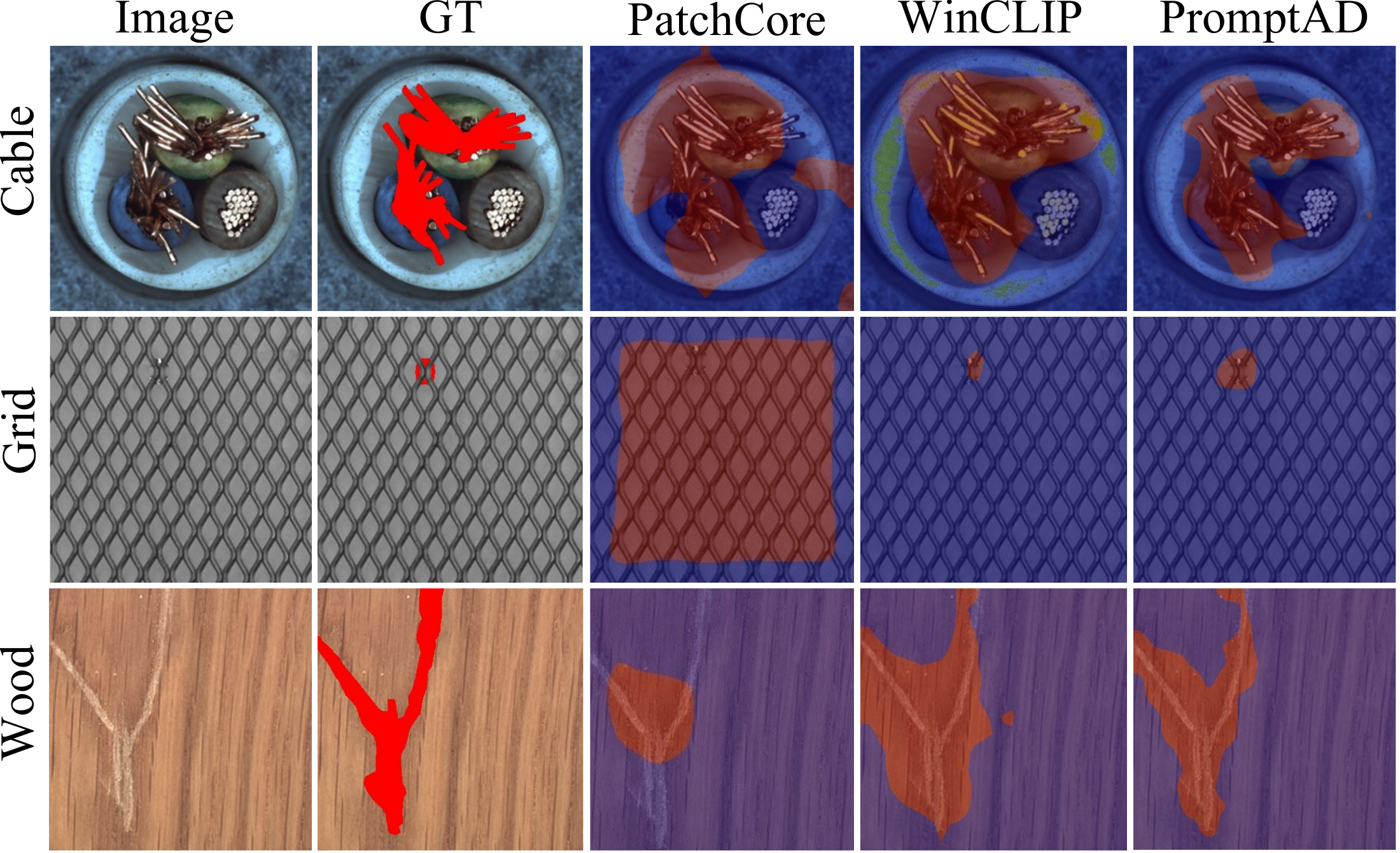}
    \caption{MVTec (1-shot)}
    \label{fig:short-a}
  \end{subfigure}
  \begin{subfigure}{0.49\linewidth}
    \includegraphics[width=1\linewidth]{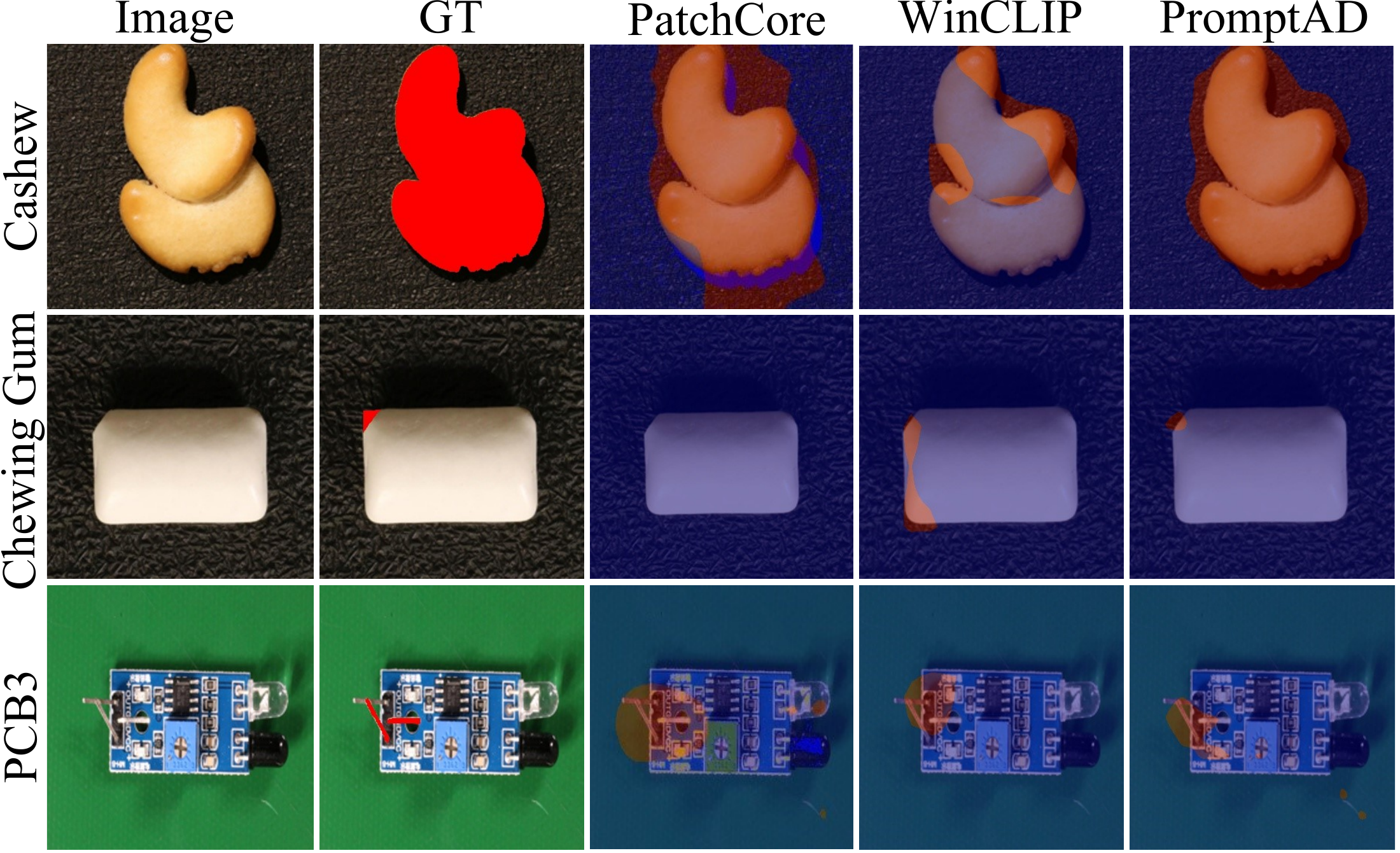}
    \caption{VisA (1-shot)}
    \label{fig:short-b}
  \end{subfigure}
  \vspace{-5pt}
  \caption{Qualitative comparison results of 1-shot pixel-level anomaly detection on MVTec \cite{MvTec} and VisA \cite{Visa}.}
  \label{fig:qua}
  \vspace{-5pt}
\end{figure*}

The quantitative results of anomaly localization are shown in Figure \ref {fig:qua}. Compared with PatchCore \cite{PatchCore} and WinCLIP+ \cite{WinCLIP}, PromptAD has a better anomaly localization capability for both objects and textures in the 1-shot setting. In addition, PromptAD can also locate some very small anomaly areas very accurately.

\subsection{Compared With Many-shot Methods}
In Table \ref {many-shot}, the comparison results of PromptAD under few-shot settings with other methods under many-shot/full-shot settings are recorded. It can be seen that compared with some methods under many-shot settings, PromptAD achieves better image-level results, and the pixel-level results are also competitive, which fully verifies the strong ability of PromptAD in the few-shot settings. In addition, PromptAD is superior to the early full-shot AD methods, MKD \cite{MKD} and P-SVDD \cite{PatchSVDD}, but there is still a certain gap between PromptAD and the latest full-shot AD methods, PatchCore \cite{PatchCore} and SimpleNet \cite{SimpleNet}.

\begin{table}[]
\centering
\scalebox{0.75}{
\setlength{\tabcolsep}{4.0mm}
\begin{tabular}{ccccccc}
\toprule
\multicolumn{2}{c}{PAD} & \multirow{2}{*}{VAD} & \multicolumn{2}{c}{MVTec} & \multicolumn{2}{c}{VisA} \\ \cmidrule(r){1-2} \cmidrule(r){4-5} \cmidrule(r){6-7}
SC & EAM & & image& pixel& image& pixel\\ \midrule
\textcolor{red}{\ding{55}} & \textcolor{red}{\ding{55}} & 
\textcolor{red}{\ding{55}} & 81.5 & 87.8 & 72.6 & 85.5 \\
\textcolor{green}{\ding{51}} & \textcolor{red}{\ding{55}} & \textcolor{red}{\ding{55}} & 90.4 & 91.7 & 81.3 & 90.5 \\
\textcolor{green}{\ding{51}} & \textcolor{green}{\ding{51}} & \textcolor{red}{\ding{55}} & \underline{91.3} & 92.5 & \underline{83.2} & 91.8 \\
\textcolor{red}{\ding{55}} & \textcolor{red}{\ding{55}} & \textcolor{green}{\ding{51}} & 85.1 & \underline{93.2} & 82.7 & \underline{95.2} \\ 
\textcolor{green}{\ding{51}} & \textcolor{green}{\ding{51}} & \textcolor{green}{\ding{51}} & \textbf{94.6} & \textbf{95.9} & \textbf{86.9} & \textbf{96.7} \\
\bottomrule
\end{tabular}
}
\caption{Image-level/pixel-level results (AUROC) of ablation study under 1-shot setting. PAD and VAD are prompt-guided and vision-guided anomaly detection, respectively, SC is semantic concatenation, and EAM is explicit anomaly margin.}
\label{ablation-study}
\vspace{-10pt}
\end{table}

\begin{table*}[]
\centering
\scalebox{0.75}{
\setlength{\tabcolsep}{7mm}
\begin{tabular}{llcccccc}
\toprule
\multicolumn{1}{l}{\multirow{2}{*}{Method}} & \multicolumn{1}{l}{\multirow{2}{*}{Public}} & \multicolumn{3}{c}{MVTec}      & \multicolumn{3}{c}{VisA} \\ \cmidrule(r){3-5} \cmidrule(r){6-8}
\multicolumn{1}{c}{} & \multicolumn{1}{c}{} & 1-shot   & 2-shot   & 4-shot   & 1-shot   & 2-shot   & 4-shot   \\ \midrule
SPADE \cite{SPADE} & arXiv'2020 & 91.2\footnotesize{$\pm$0.4} & 92.0\footnotesize{$\pm$0.3} & 92.7\footnotesize{$\pm$0.3} & 95.6\footnotesize{$\pm$0.4} & 96.2\footnotesize{$\pm$0.4} & 96.6\footnotesize{$\pm$0.3} \\
PaDiM \cite{PaDiM} & ICPR'2020 & 89.3\footnotesize{$\pm$0.9} & 91.3\footnotesize{$\pm$0.7} & 92.6\footnotesize{$\pm$0.7} & 89.9\footnotesize{$\pm$0.8} & 92.0\footnotesize{$\pm$0.7} & 93.2\footnotesize{$\pm$0.5} \\
PatchCore \cite{PatchCore} & CVPR'2022 & 92.0\footnotesize{$\pm$1.0} & 93.3\footnotesize{$\pm$0.6} & 94.3\footnotesize{$\pm$0.5} & 95.4\footnotesize{$\pm$0.6} & 96.1\footnotesize{$\pm$0.5} & 96.8\footnotesize{$\pm$0.3} \\
WinCLIP+\dag\ \cite{WinCLIP} & CVPR'2023 & \underline{95.2\footnotesize{$\pm$0.5}} & \underline{96.0\footnotesize{$\pm$0.3}} & 96.2\footnotesize{$\pm$0.3} & \underline{96.4\footnotesize{$\pm$0.4}} & \underline{96.8\footnotesize{$\pm$0.3}} & \underline{97.2\footnotesize{$\pm$0.2}} \\
FastRecon \cite{FastRecon} & ICCV'2023                                   & - & 95.9 & \textbf{97.0} & - & - & - \\ \midrule
\multicolumn{1}{l}{\textbf{PromptAD}\dag}                     & \multicolumn{1}{c}{-} & \textbf{95.9\footnotesize{$\pm$0.5}} & \textbf{96.2\footnotesize{$\pm$0.3}} & \underline{96.5\footnotesize{$\pm$0.2}} & \textbf{96.7\footnotesize{$\pm$0.4}} & \textbf{97.1\footnotesize{$\pm$0.3}} & \textbf{97.4\footnotesize{$\pm$0.3}} \\ \bottomrule
\end{tabular}
}
\caption{Comparison of pixel-level anomaly detection in AUROC on MVTec and VisA benchmarks. The best and second-best results are respectively marked in \textbf{bold} and underlined. $\dag$ indicates CLIP-based methods.}
\vspace{-10pt}
\label{pixel-level}
\end{table*}

\subsection{Ablation Study}
We verify the impact of different modules of different proposed methods on the overall performance of PromptAD under 1-shot setting on MVTec \cite{MvTec} and VisA \cite{Visa}. These include semantic concatenation (SC) and explicit anomaly margin (EMA). Meanwhile, we also verified the effect of vision-guided anomaly detection (VAD). Results of the ablation study are recorded in Table \ref {ablation-study}.

\vspace{0.5\baselineskip}
\noindent \textbf{Semantic Concatenation (SC).} 
The number of negative samples plays a crucial role in contrastive learning \cite{SimCLR, MOCO}. Without the proposed SC, the conventional prompt learning paradigm \cite{COOP} loses negative prompts for contrast, so the effect of prompt learning will be greatly reduced. As shown in Table \ref {ablation-study}, there is a significant drop in image and pixel level results on both MVTec \cite{MvTec} and VisA \cite{Visa} when SC is not used. After using SC, the image-level/pixel-level results on MVTec (and VisA) are improved by 8.9\%/3.9\% (8.9\%/5.0\%), which indicates that SC can greatly improve the applicability of prompt learning in anomaly detection.

\vspace{0.5\baselineskip}
\noindent \textbf{Explicit Anomaly Margin (EAM).} Since anomaly samples are absent during the training phase, it is hard to establish an explicit margin between the features of normal and anomaly prompts. EAM uses a hyper-parameter to control the margin between normal and anomaly prompt features, which can make up for the lack of contrastive loss. Table \ref {ablation-study} shows that after using EAM, the image-level/pixel-level results on MVTec (and VisA) are improved by 0.9\%/0.8\% (1.9\%/0.7\%), respectively. 

\vspace{0.5\baselineskip}
\noindent \textbf{Vision-guided Anomaly Detection (VAD).} PAD introduces more high-level semantic information but ignores many local details, which is not conducive to pixel-level anomaly detection. On the contrary, VAD using normal feature memory focuses on more local detail information. In Table \ref {ablation-study}, PAD has better image-level results, while VAD has better pixel-level results, and the two have a good complementarity. Under the 1-shot setting, the results of PAD and VAD are fused by harmonic mean, and 94.6\%/95.9\% (86.8\%/96.7\%) image-level/pixel-level results are achieved on MVTec (and VisA).

\begin{table}[]
\centering
\scalebox{0.75}{
\setlength{\tabcolsep}{8.0mm}
\begin{tabular}{lcc}
\toprule
 Method & MVTec & VisA \\ \midrule

CLIP \cite{CLIP} & 22.5 & 24.6 \\
CLIP \cite{CLIP} + ours & 79.9 & 80.4 \\
CLIP+Linear \cite{VAND} + ours & 79.4 & 77.2 \\ \midrule
MaskCLIP \cite{MaskCLIP} & 85.5 & 80.5 \\
MaskCLIP \cite{MaskCLIP} + ours & \underline{91.6} & \underline{91.2} \\ \midrule
VV-CLIP \cite{CLIPSurgery} & 86.7 & 82.9 \\
VV-CLIP \cite{CLIPSurgery} + ours & \textbf{92.5} & \textbf{91.8} \\ 

\bottomrule
\end{tabular}
}
\caption{Pixel-level results (AUROC) of using different CLIP transformations on MVTec and VisA under 0-shot/1-shot settings.}
\label{CLIP-Trans}
\vspace{-14pt}
\end{table}

\subsection{Results of Different CLIP Transformations}
Due to the inability of CLIP to directly complete prompt-guided localization tasks, some works have explored the transformations of CLIP \cite{CLIPSurgery, MaskCLIP}. Table \ref {CLIP-Trans} records the results of different CLIP transformations under pixel-level anomaly detection, where about 1000 prompts are used in 0-shot setting and our prompt learning method is used in 1-shot setting. MaskCLIP \cite{MaskCLIP} drops the QKV attention and leaves only $V\_Proj.$ and $Proj.$, and then embed local features after each layer of the visual encoder as in VV-CLIP. CLIP+Linear \cite{VAND} adds a learnable linear layer to the visual encoder after each block to align the local features with prompt features.

As shown in Table \ref {CLIP-Trans}, the results of the original CLIP under 0-shot are 22.5\%/24.6\% on MVTec/Visa, which is lower than the random prediction (50.0\%). This is caused by the opposite visual activation \cite{CLIPSurgery, ECLIP} of CLIP. After the transformation of attention, MaskCLIP \cite{MaskCLIP} and VV-CLIP \cite{CLIPSurgery} achieve a huge improvement of 63.0\%/55.9\% and 64.2\%/58.3\% on MVTec/Visa, respectively. The improvement of VV-CLIP  is more obvious than that of MaskCLIP. We speculate that this is because VV-attention retains a certain information interaction while focusing on local information, while MaskCLIP completely removes attention.

After using our method, the pixel-level results of MaskCLIP and VV-CLIP are increased by 6.1\%/10.7\%, and 5.8\%/9.8\% on MVTec/Visa, respectively. Furthermore, it is worth noting that prompt learning also leads to a significant 57.4\%/55.8\% improvement in pixel-level results of the original CLIP. However, when prompt learning is added with learnable linear layers, the effect decreases, which may be because there is mutual interference between prompt learning and the training of linear layers.

\begin{figure}
  \centering
  \begin{subfigure}{0.4\linewidth}
    \includegraphics[width=1\linewidth]{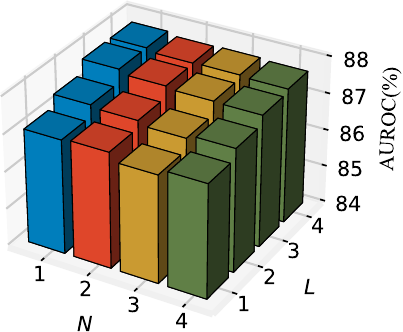}
    \caption{Image-level results.}
    \label{fig:short-a}
  \end{subfigure}
  \quad
  \begin{subfigure}{0.4\linewidth}
    \includegraphics[width=1\linewidth]{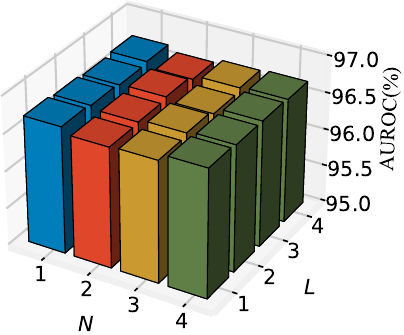}
    \caption{Pixel-level results.}
    \label{fig:short-b}
  \end{subfigure}
  \vspace{-5pt}
  \caption{Image-level/pixel-level results on VisA \cite{Visa} in 1-shot setting using different $N$ and $L$.}
  \label{fig:NL}
\end{figure}

\begin{figure}[]
  \centering
   \includegraphics[width=0.7\linewidth]{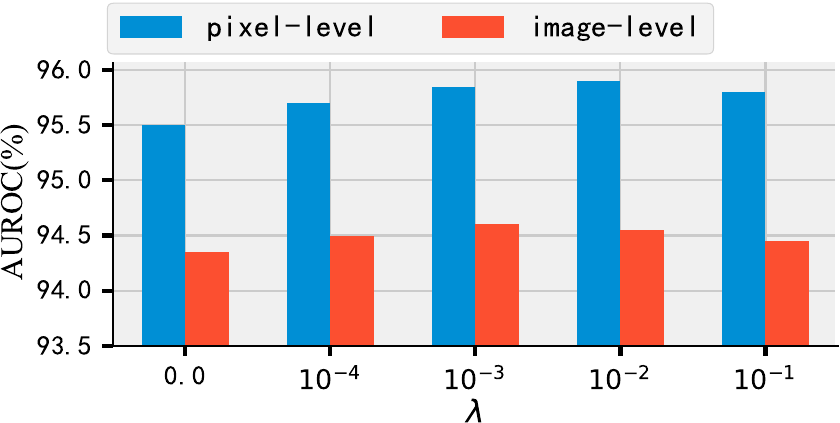}

   \caption{Image-level/pixel-level results on MVTec \cite{MvTec} in the 1-shot setting using different hyper-parameter $\lambda$.}
   \vspace{-12pt}
   \label{fig:lambda}
\end{figure}

\subsection{Hyper-parameter Analysis}

We complete the effect of $N$, $L$ and $\lambda$ on PromptAD. $\lambda$ is the hyper-parameter of the loss $\mathcal{L}_{algin}$, which controls the degree of alignment between MAPs and LAPs feature distributions. $N$ is the number of NPs. $L$ is the number of anomaly prompt suffixes, and $N\times L$ is the number of LAPs.

Figure \ref {fig:NL} illustrates the effect of, $N$ and $L$ on PromptAD. In Image-level results, $N$ does not have a great influence, and there is no significant difference between $N=1$ and $N=4$. While, $L$ has a significant influence, and larger $L$ can lead to higher results. In pixel-level results, the effects of both $N$ and $L$ are relatively small, and larger $L$ slightly improves the results. Figure \ref {fig:lambda} records the image-level/pixel-level results with different $\lambda$. It can be seen that the results are worse when the $\lambda$ is equal to 0 or larger. This indicates that the distributions of MAPs and LAPs need to be aligned, but not over-aligned, which will reduce the diversity of the anomaly prompts and thus reduce the model's perception of anomaly image features. 

\subsection{Visualization Results}

\begin{figure}
  \centering
  \begin{subfigure}{0.4\linewidth}
    \includegraphics[width=1\linewidth]{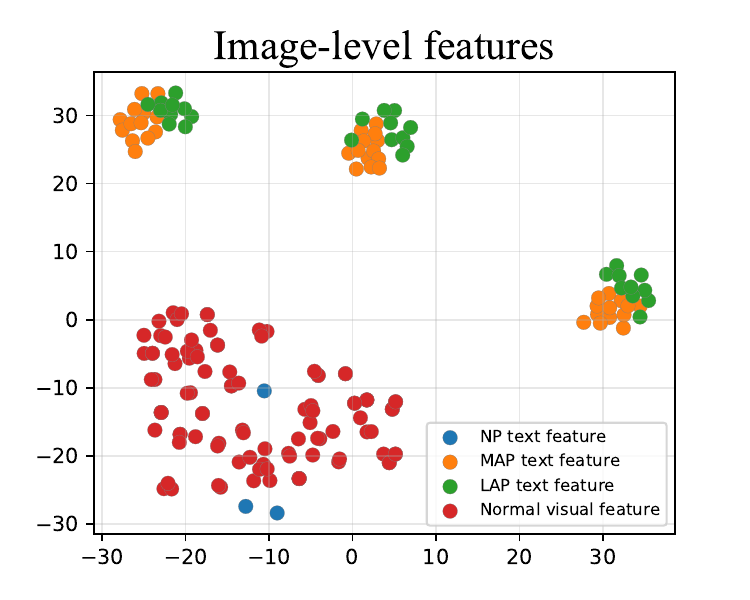}
    \label{fig:short-a}
  \end{subfigure}
  \ \ \ \ \
  \begin{subfigure}{0.4\linewidth}
    \includegraphics[width=1\linewidth]{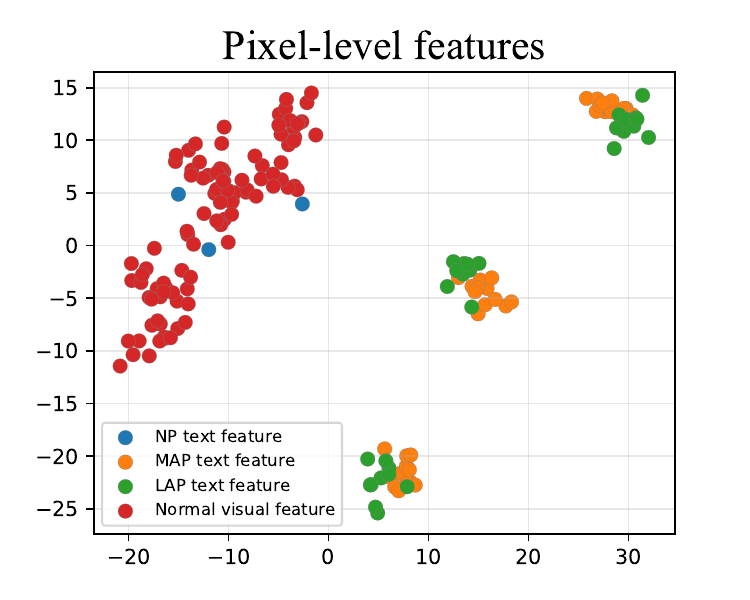}
    \label{fig:short-b}
  \end{subfigure}
  \vspace{-5pt}
  \caption{Feature visualization results using T-SNE in 1-shot setting. The feature used is ``cable" from the MVTec \cite{MvTec}.}
  \vspace{-10pt}
  \label{fig:prompt_tsne}
\end{figure}

To quantify the effect of PromptAD, we visualize the visual and textual features after $L_2$ normalization. Specifically, we visualize 3 NPs, 3$\times$13 MAPs, and 3$\times$10 LAPs as well as 100 image-level/pixel-level normal visual features. Figure \ref {fig:prompt_tsne} shows the visualization results, it can be seen that
there is very clear discrimination between normal prompt features and anomaly prompt features, and the overlap between normal prompt features and normal visual features
is very high, which intuitively verifies the effectiveness of PromptAD. In addition, it’s worth noting that the 3 normal prompt features do not collapse to one point, but fit the overall distribution of normal visual features as much as possible.

\section{Conclusion}
In this paper, we propose a novel anomaly detection method termed PromptAD which automatically learns prompts with only normal samples in the few-shot anomaly detection scenario. First, in order to cope with the challenge under the one-class task, we propose semantic concatenation to construct enough anomaly prompts through concatenating normal prompts and anomaly suffixes to guide prompt learning. Second, we propose the explicit anomaly margin loss, which explicitly determines the margin between normal prompt features and anomaly prompt features through a hyper-parameter. Finally, for image-level/pixel-level anomaly detection, PromptAD achieves first place in
11/12 few-shot tasks.

\noindent\textbf{Acknowledgments}
This work is supported by the National Natural Science Foundation of China No. 62176092, 62222602, U23A20343, 62106075, 62302167, Shanghai Sailing Program (23YF1410500), Natural Science Foundation of Shanghai (23ZR1420400),  Science and Technology Commission of Shanghai No.21511100700, Natural Science Foundation of Chongqing, China (CSTB2023NSCQ-JQX0007, CSTB2023NSCQ-MSX0137), CCF-Tencent Rhino-Bird Young Faculty Open Research Fund (RAGR20230121), China Postdoctoral Science Foundation funded project (2023M734270), and Development Project of Ministry of Industry and Information Technology (ZTZB.23-990-016).

{
    \small
    \bibliographystyle{ieeenat_fullname}
    \bibliography{main}
}
\clearpage
\maketitlesupplementary
\renewcommand\thesection{\Alph{section}}

\setcounter{page}{1}
\setcounter{section}{0}

\begin{figure*}[hb!]
  \centering
   \includegraphics[width=1.0\linewidth]{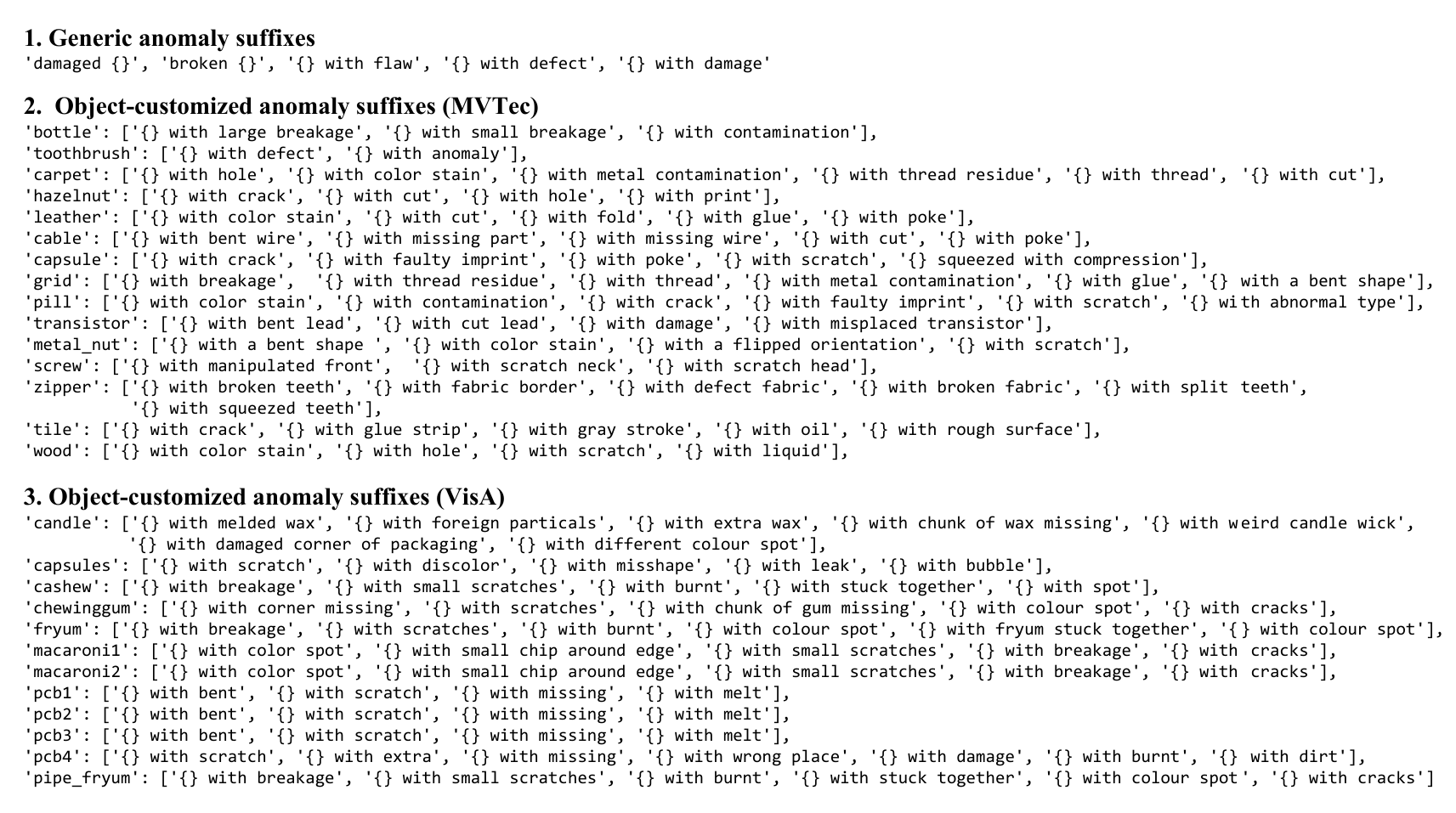}
    
   \caption{Illustration of manual anomaly suffixes.}
   \label{MAP}
\end{figure*}

\section{Experimental details}
\label{sec:details}
\noindent \textbf{Data pre-processing}. Referring to WinCLIP \cite{WinCLIP}, we employ the data pre-processing pipeline specified in OpenCLIP \cite{OpenCLIP} for both the MVTec \cite{MvTec} and VisA \cite{Visa} datasets to mitigate potential train-test discrepancies. This involves channel-wise standardization using precomputed mean [0.48145466, 0.4578275, 0.40821073] and standard deviation [0.26862954, 0.26130258, 0.27577711] after normalizing each RGB image to [0, 1]. Subsequently, bicubic resizing is performed based on the \verb'Pillow' implementation. As a default, we set the input resolution to 240 for the shorter edge resulting from resizing, aligning with ViT-B/16$\pm$ in our experiments. 

\noindent \textbf{Hyper-parameter.} The length of trainable tokens in normal prompts ($E_N$) is set to $4$, and the length of trainable tokens in learnable anomaly prompts ($E_A$) is set to $1$. For each detection object, the number of normal prompts ($N$) is set to $1$, the number of learnable anomaly suffixes ($L$) is set to $4$, and the number of manual anomaly suffixes ($M$) depends on the number of anomaly labels in the dataset. $\lambda$ is set to $0.001$. Referring to CoOp \cite{COOP}, the optimizer parameters of prompt learning: learning rate, momentum, and weight decay are set to $0.002$, $0.9$, and $0.0005$, respectively.

\noindent \textbf{Manual anomaly suffixes.} We used two kinds of manual anomaly suffixes: generic anomaly suffixes and object-customized anomaly suffixes. Generic anomaly suffixes are manually designed and object-customized anomaly suffixes are generated through anomaly labels in the datasets \cite{MvTec, Visa}. The specific details are shown in Figure \ref{MAP}.

\noindent \textbf{Evaluation metrics.} In addition to the results for the Area Under the Receiver Operator Curve documented in the body of the paper, We also supplement the image-level Precision-Recall (AUPR) results and pixel-level Per-region-overlap (PRO) \cite{MvTec, UniS-T} results.

\noindent \textbf{Other details.} Since model performance in the few-shot setting is affected by random sampling, we report the mean and standard deviation over 5 random seeds for each measurement. In addition, the few-shot results of SPADE \cite{SPADE}, PaDiM \cite{PaDiM}, and PatchCore \cite{PatchCore} in the experiments adopt the results recorded in WinCLIP \cite{WinCLIP}.

\begin{figure*}[t!]
  \centering
   \includegraphics[width=0.95\linewidth]{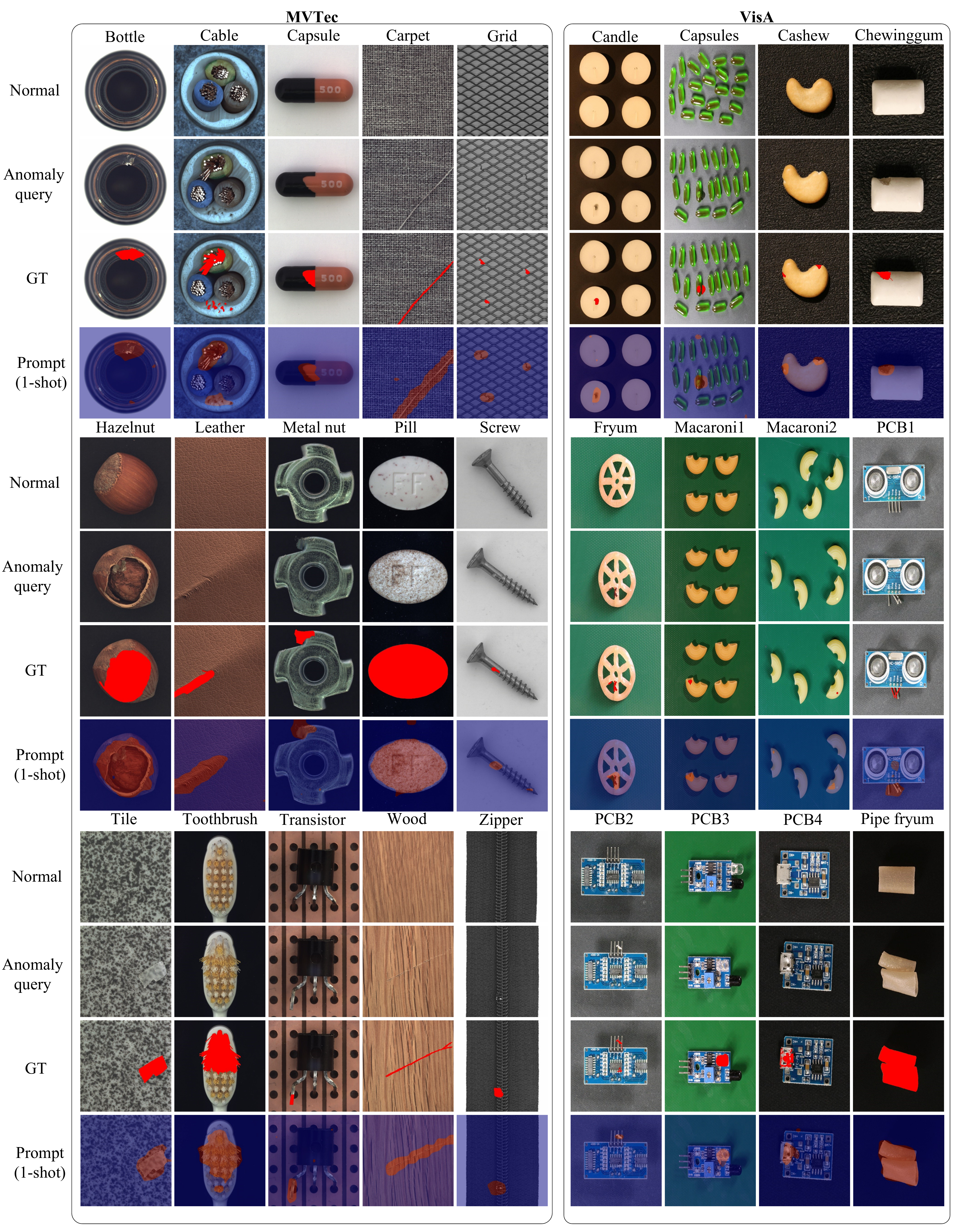}
   \caption{Additional qualitative results from PromptAD (1-shot), tested on MVTec \cite{MvTec} and VisA \cite{Visa}.}
   \label{add_quantitative}
\end{figure*}


\begin{figure}[t!]
  \centering
   \includegraphics[width=1\linewidth]{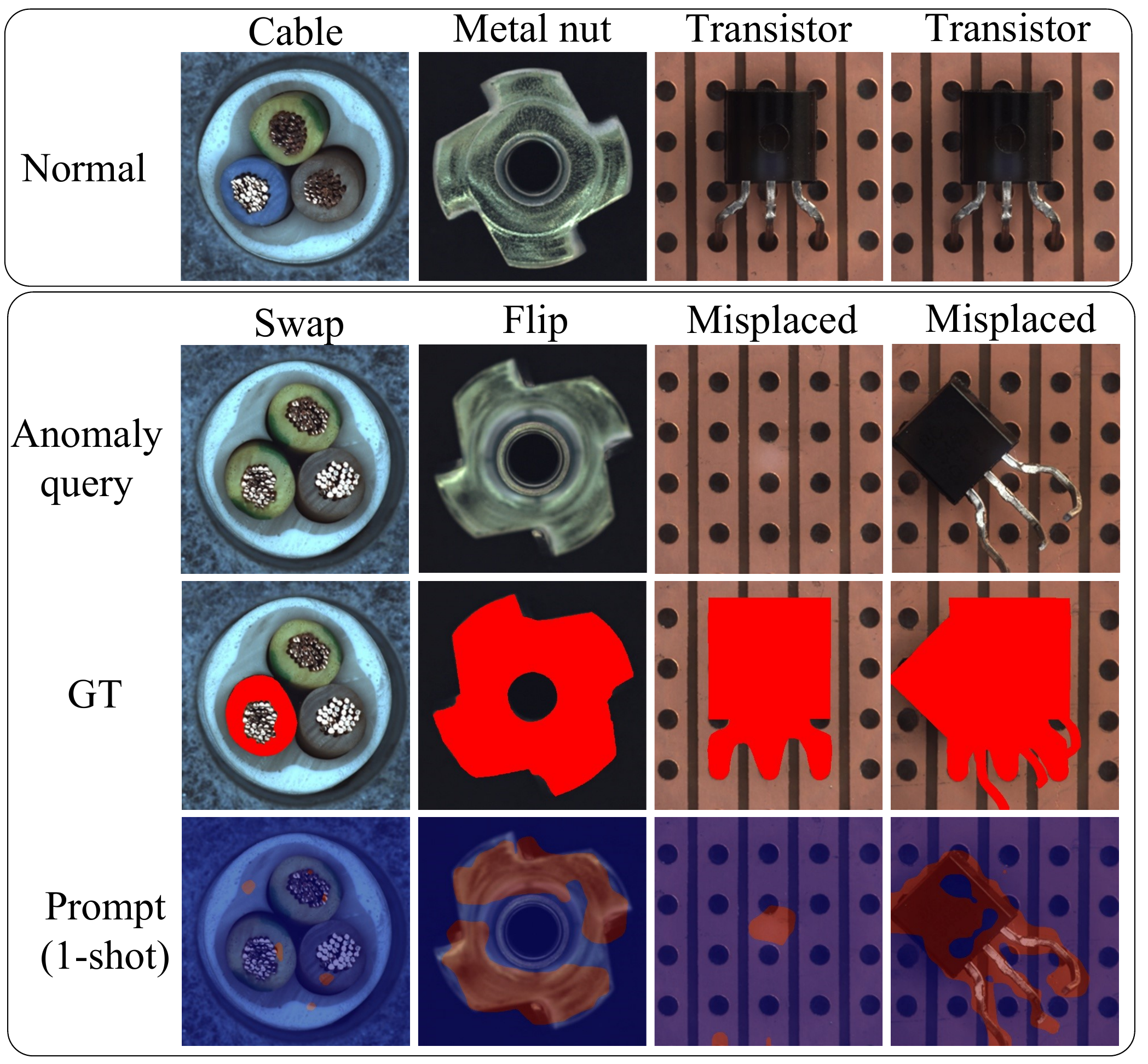}
   \caption{Qualitative results of logical anomaly detection.}
   \label{logi}

\end{figure}

\section{Additional Qualitative Results}
\label{sec:qualitative}
\begin{figure}[t!]
  \centering
   \includegraphics[width=1\linewidth]{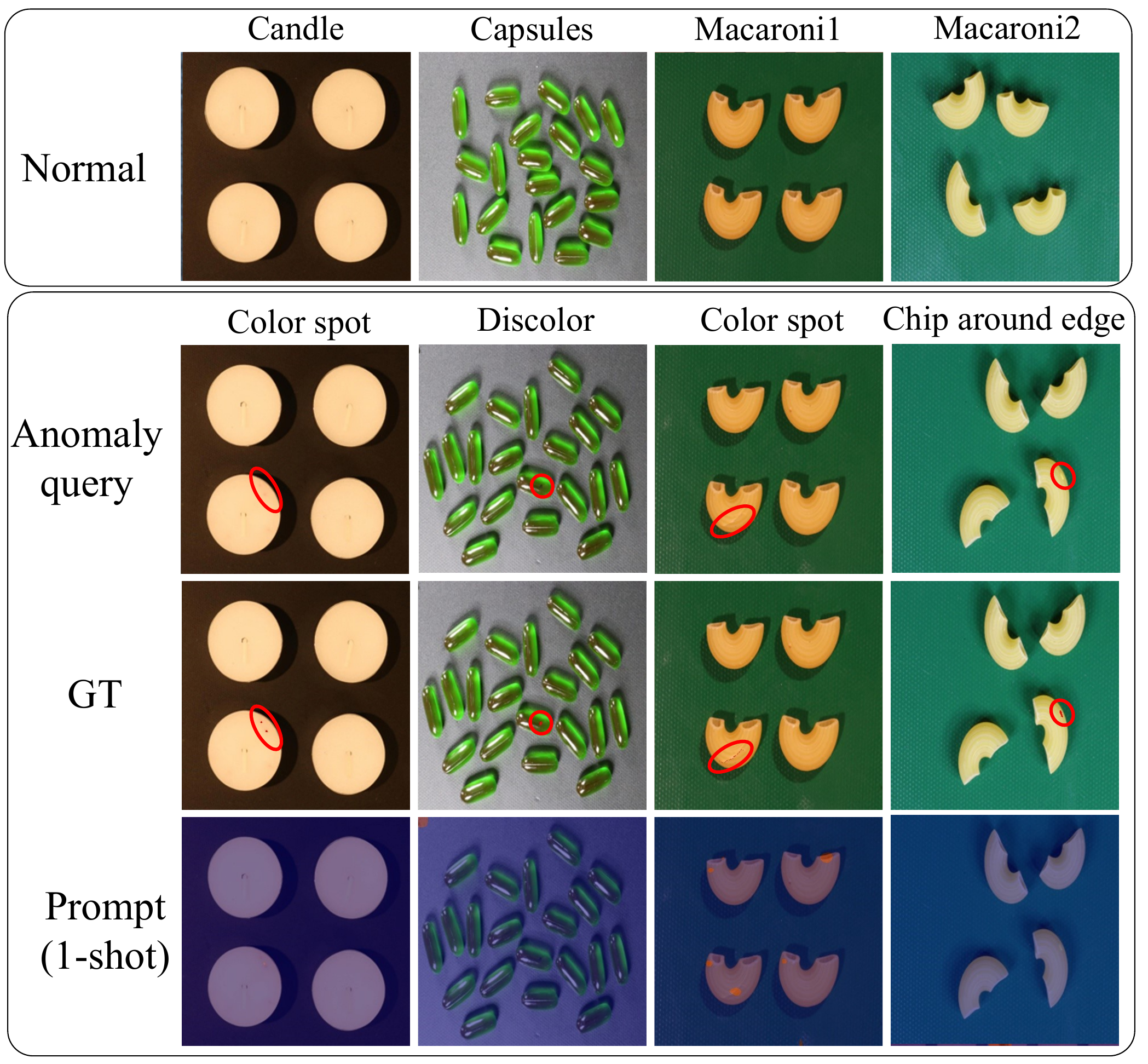}
   \caption{Qualitative results of extremely small anomaly detection.}
   \label{small}

\end{figure}

In Figure \ref{add_quantitative}, we provide further qualitative results obtained from our (1-shot) PromtAD
for pixel-level anomaly detection in MVTec \cite{MvTec} and VisA \cite{Visa}. It can be seen that PromptAD can accurately locate both large-area surface defects and small-area surface defects. In addition, as shown in Figure \ref{logi}, we also provide quantitative results of PromptAD on some logical anomalies. Logical anomalies are mainly found in some industrial components in MVTec. It can be seen that PromptAD has poor detection results on the swap anomaly of ``cable'' and missing anomaly of ``transistor'', while PromptAD has good localization results on the flip anomaly of ``Metal nut'' and misplaced anomaly of ``Transistor''. Figure \ref{small} shows PromptAD's detection results for extremely small anomalies, which are usually hard to detect by humans. For the convenience of viewing, we circle the anomaly positions in red circles, and it can be seen that PromptAD has difficulty in completing the accurate localization of extremely small anomalies.

\section{Comparison with Other Prompt Learning Methods}
\label{sec:pl}

PromptAD is the first prompt design paradigm for one class classification (OCC) promblem, which overcomes the poor performance of other prompt learning methods in anomaly detection. See Table \ref{table1}, the classical prompt learning methods (1-shot) perform relatively poorly in anomaly detection tasks, and except for the pixel-level results of CoOp \cite{COOP} on MVTec and VisA, the results of CoOp \cite{COOP}, CoCoOp \cite{COCOOP} and Maple \cite{MaPLe} are even worse than contextual prompt engineering (CPE). PromptAD not only outperforms the classical prompt learning method in anomaly detection tasks, but also, compared with CPE, brings improvements of 0.8\%/5.8\% and 5.0\%/8.9\% on MVTec and VisA respectively (image-level/pixel-level).

\begin{table}[h]
\centering
\scalebox{0.9}{
\begin{tabular}{cccccc}
\toprule
& \multirow{2}{*}{Method} & \multicolumn{2}{c}{MVTec} & \multicolumn{2}{c}{VisA} \\ \cmidrule(r){3-6}
& & image& pixel& image& pixel\\ \midrule
 
\multirow{2}{*}{\rotatebox{90}{0-shot}}& CPE+WinCLIP & 91.8 & 85.1 & 78.1 & 79.6 \\

&CPE+VV-CLIP & 90.5 & 86.7 & 77.2 & 82.9 \\
\cmidrule(r){1-6}

&CoOp+VV-CLIP & 81.5 & 87.8 & 72.6 & 85.5 \\

\multirow{2}{*}{\rotatebox{90}{1-shot}}&CoCoOp+VV-CLIP & 60.6 & 52.1 & 61.6 & 72.3 \\

&Maple & 66.8 & 64.9 & 60.5 & 61.5 \\
\cmidrule(r){2-6}
&PromptAD & \textbf{91.3} & \textbf{92.5} & \textbf{83.2} & \textbf{91.8} \\

\bottomrule
\end{tabular}
}
\caption{Results (AUROC) of different methods (w/o VAD).}
\label{table1}
\end{table}

\section{Ablation Study}
\label{sec:lm}

We further evaluate the impact of MAPs and LAPs on PromptAD, respectively. For better comparison, we removed the effect of vision-guided anomaly detection (VAD). Table \ref{table2} (second row) shows that without MAPs, the performance degradation is more pronounced, but it is still better than CPE, indicating that PromptAD can still automatically learn effective anomaly prompts even without using any manually annotated anomaly description suffix. Table \ref{table2} (third row) shows that there is also a slight decrease in performance without LAPs, which indicates that LAPs can expand the set of exception hints and thus improve the model performance.

\begin{table}[]
\centering
\scalebox{0.9}{
\begin{tabular}{cccccc}
\toprule
\multirow{2}{*}{} & \multirow{2}{*}{Method} & \multicolumn{2}{c}{MVTec} & \multicolumn{2}{c}{VisA} \\ \cmidrule(r){3-6}
&  & image& pixel& image& pixel\\ \midrule
1 & CPE+VV-CLIP & 90.5 & 86.7 & 77.2 & 82.9 \\

2 & Prompt w/o MAPs & 87.6 & 89.8 & 78.8 & 86.2 \\

3 & Prompt w/o LAPs & 90.5 & 91.3 & 82.5 & 90.3 \\

4 & PromptAD & \textbf{91.3} & \textbf{92.5} & \textbf{83.2} & \textbf{91.8} \\

\bottomrule
\end{tabular}
}

\caption{Effects (AUROC) of MAPs and LAPs (1-shot \& w/o VAD).}
\label{table2}
\end{table}

In addition, we explore the impact of CLIP's different visual backbones on PromptAD. The results of using different backbones are recorded in Table \ref{table4}, where the self-attention modules of VIT add the vv-attention branches, and the attention pooling of ResNet101 adopts V-V attention. All of the backbones require no additional training. The overall performance of ViT is better than ResNet, VIT-L/14 shows better pixel-wise anomaly detection than VIT-B/16+.

\begin{table}[]
\centering
\scalebox{0.90}{
\setlength{\tabcolsep}{2.0mm}
\begin{tabular}{ccccc}
\toprule
\multirow{1}{*}{Backbone} & Dataset & 1-shot& 2-shot& 4-shot\\ \midrule

ResNet101 & MVTec & 85.8/93.0 & 87.6/94.3 & 90.3/94.8 \\

ViT-L/14 & MVTec & 92.4/95.5 & 93.8/95.8 & 94.9/96.2 \\

ViT-B/16+ & MVTec & \textbf{94.6}/\textbf{95.9} & \textbf{95.7}/\textbf{96.2} & \textbf{96.6}/\textbf{96.5} \\ \midrule

ResNet101 & VisA & 80.4/95.1 & 84.5/96.3 & 85.3/96.9 \\

ViT-L/14 & VisA & 85.2/\textbf{96.8} & 86.3/\textbf{97.2} & 86.7/\textbf{97.4} \\

ViT-B/16+ & VisA & \textbf{86.9}/96.7 & \textbf{88.3}/97.1 & \textbf{89.1}/\textbf{97.4} \\ 

\bottomrule

\end{tabular}
}

\caption{Image-level/pixel-level results (AUROC) with other backbones.}
\label{table4}
\end{table}

\section{Results on Other Benchmarks}
In addition to the two datasets MVTec \cite{MvTec} and VisA \cite{Visa}, we also evaluate PromptAD in the few-shot setting on MPDD \cite{MPDD} and LOCO \cite{LOCO}. We reproduce the results of PatchCore and WinCLIP+. As shown in Table \ref{table3}, compared with PatchCore and WinCLIP+, PromptAD achieves the first place in few-shot settings of both datasets

\begin{table}[]
\centering
\scalebox{0.9}{
\begin{tabular}{ccccc}
\toprule
\multirow{1}{*}{Method} & Dataset & 1-shot& 2-shot& 4-shot\\ \midrule

PatchCore & MPDD & 59.2/78.5 & 59.6/79.2 & 79.9/79.8 \\

WinCLIP+ & MPDD & 68.2/92.6 & 69.3/94.7 & 75.2/96.0 \\

PromptAD & MPDD & \textbf{80.7}/\textbf{96.2} & \textbf{85.3}/\textbf{97.2} & \textbf{87.2}/\textbf{97.3} \\ \midrule

PatchCore & LOCO & 64.9/70.3 & 65.4/71.5 & 68.7/72.2 \\

WinCLIP+ & LOCO & 68.0/71.2 & 69.7/71.9 & 71.3/72.8 \\

PromptAD & LOCO & \textbf{71.2}/\textbf{73.0} & \textbf{72.6}/\textbf{74.1} & \textbf{73.5}/\textbf{74.5} \\

\bottomrule

\end{tabular}
}
\caption{Image-level/pixel-level results (AUROC) on MPDD and LOCO.}
\label{table3}
\end{table}

\section{Detailed Comparison Results}
\label{sec:detailed-comparison}

In this section, we report the detailed subset-level results of PromptAD. In addition, we evaluate PromptAD's results on Image-level AUPR and pixel-level PRO. Specifically, the results for MVTec are recorded in Table \ref{MVTec-I-ROC},\ref{MVTec-P-ROC},\ref{MVTec-I-PR},\ref{MVTec-P-PRO} and the results for VisA are recorded in Table \ref{VisA-I-ROC},\ref{VisA-P-ROC},\ref{VisA-I-PR},\ref{VisA-P-PRO} with the first place marked in bold and the second place mean underlined.

\section{Visualization Results of Attention Map }
\label{sec:vram}

\begin{figure*}[]
  \centering
   \includegraphics[width=1.0\linewidth]{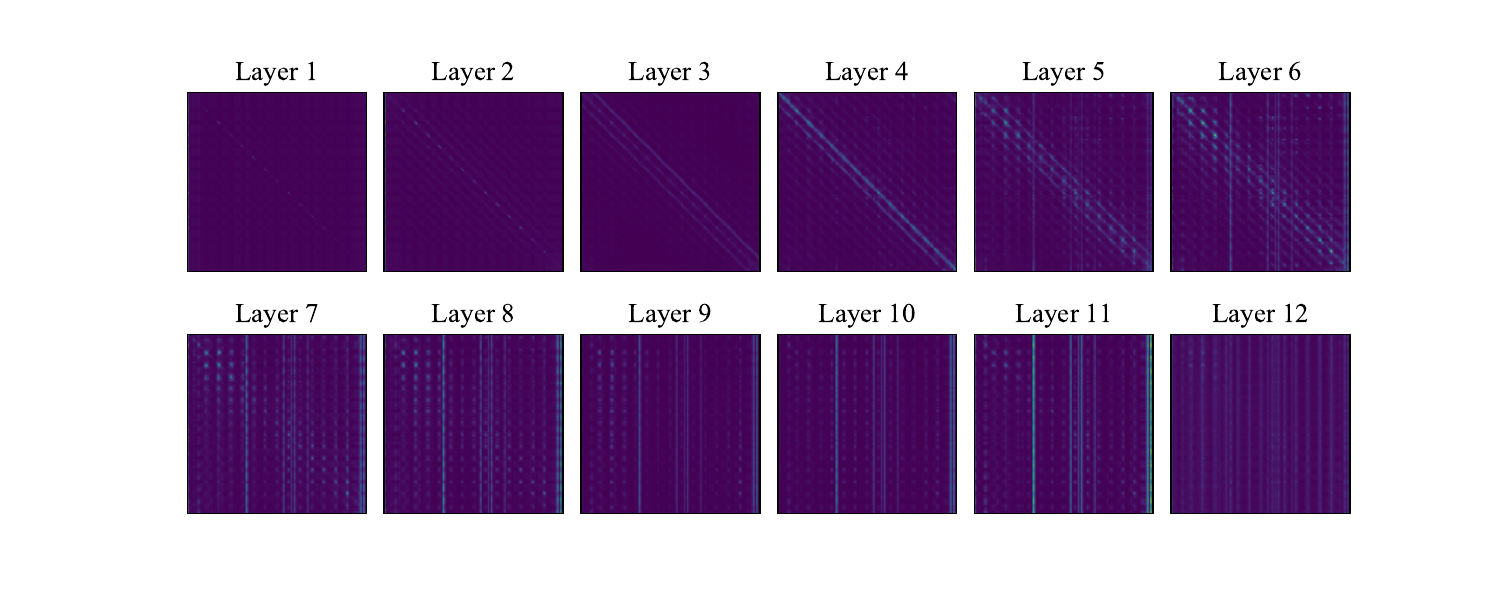}
   \caption{Visualization of the QK attention map in the vision encoder.}
   \label{QK}

  \centering
   \includegraphics[width=1.0\linewidth]{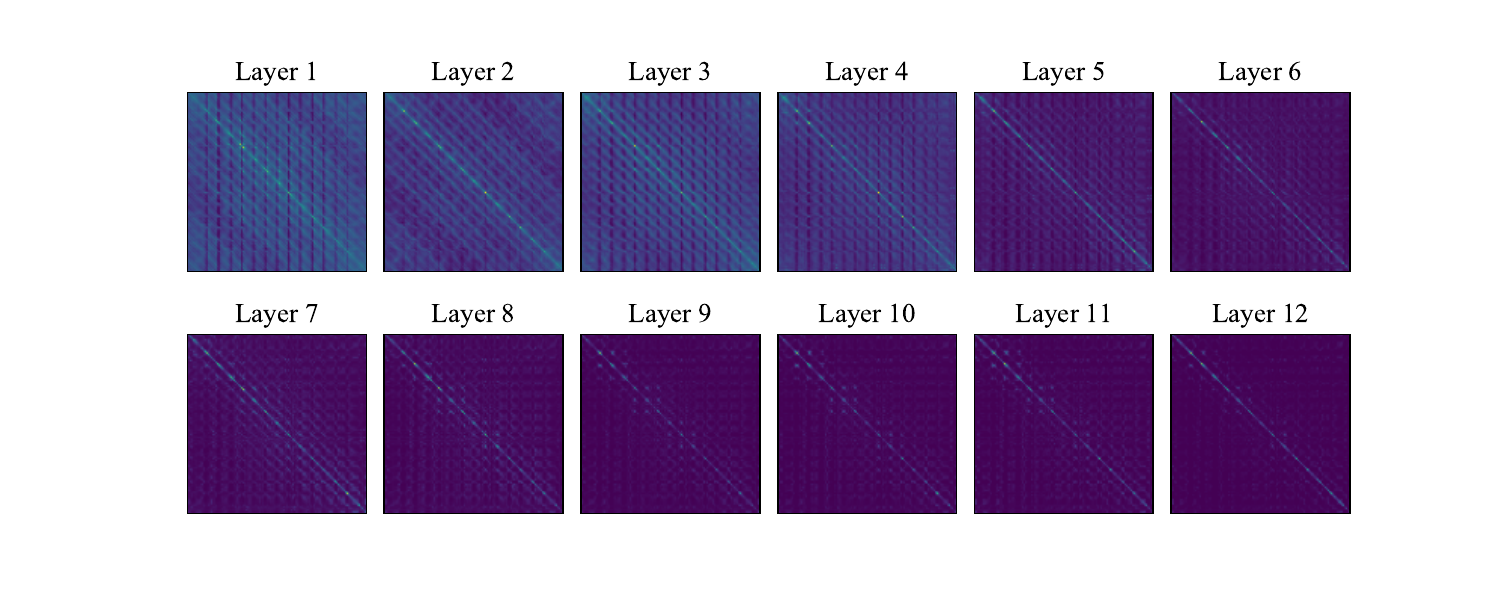}
   \caption{Visualization of the VV attention map in the vision encoder.}
   \label{VV}
\end{figure*}

To further analyze the working mechanism of CLIP \cite{CLIP}, we provide visualization results of the attention map in the CLIP visual encoder (ViT-B/16+). Figure \ref{QK} is the visualization result of the original QK attention map. It can be seen that QK attention in the shallow layers focuses more on local information (main diagonal activations). From layer 5, QK attention starts to focus on more global information. Both global and local information play an important role in anomaly detection. Therefore, to preserve both local and global information, PromptAD uses the features of the $3^{th}$ and $8^{th}$ layers instead of the features of the $6^{th}$ and $9^{th}$ layers used by PatchCore \cite{PatchCore} when storing visual features. The comparison results are reported in Table \ref{layer}, compared with $6^{th}$+$9^{th}$ features, there is a 1.2\% improvement using the $3^{th}$+$8^{th}$ features. Figure \ref{VV} shows the visualization results of the VV attention map, it can be seen that compared with QK attention, VV attention focuses on local information from the first layer to the last layer, which is more conducive to completing the localization task. While, most layers of QK attention focus on global information, which is more conducive to classification tasks.

\begin{table}[]
\centering
\begin{tabular}{c|ccc|c}
\toprule
          & screw & grid & transistor & mean \\ \midrule
$6^{th}$+$9^{th}$ & 91.7  & 58.8 & 84.9       & 92.0   \\
$3^{th}$+$8^{th}$ & 89.6  & 81.8 & 87.2       & 93.2 \\ \bottomrule
\end{tabular}
\caption{Pixel-level results (AUROC) of using different layer features as the feature memory.}
\label{layer}
\end{table}

\begin{table*}[]
\centering
\resizebox{\linewidth}{!}{
\begin{tabular}{c|ccccc|ccccc|ccccc}
\toprule
MVTec & \multicolumn{5}{c|}{1-shot} & \multicolumn{5}{c|}{2-shot} & \multicolumn{5}{c}{4-shot} \\ 
Image-AUROC & SPADE     & PaDiM    & PatchCore & WinCLIP+  & \cellcolor[HTML]{EFEFEF}PromptAD  & SPADE     & PaDiM    & PatchCore & WinCLIP+  & \cellcolor[HTML]{EFEFEF}PromptAD  & SPADE     & PaDiM    & PatchCore & WinCLIP+   & \cellcolor[HTML]{EFEFEF}PromptAD  \\ \midrule
Bottle      & 98.7$\pm$0.6  & 97.4$\pm$0.7 & 99.4$\pm$0.4  & 98.2$\pm$0.9 & \cellcolor[HTML]{EFEFEF}99.8$\pm$0.3  & 99.5$\pm$0.1  & 98.5$\pm$1.0 & 99.2$\pm$0.3  & 99.3$\pm$0.3 & \cellcolor[HTML]{EFEFEF}99.4$\pm$0.1 & 99.5$\pm$0.2  & 98.8$\pm$0.2 & 99.2$\pm$0.3  & 99.3$\pm$0.4  & \cellcolor[HTML]{EFEFEF}99.8$\pm$0.2 \\
Cable       & 71.2$\pm$3.3  & 57.7$\pm$4.6 & 88.8$\pm$4.2  & 88.9$\pm$1.9 & \cellcolor[HTML]{EFEFEF}94.2$\pm$1.2  & 76.2$\pm$5.2  & 62.3$\pm$5.9 & 91.0$\pm$2.7  & 88.4$\pm$0.7 & \cellcolor[HTML]{EFEFEF}91.8$\pm$1.5  & 83.4$\pm$3.1  & 70.0$\pm$6.1 & 91.0$\pm$2.7  & 90.9$\pm$0.9  & \cellcolor[HTML]{EFEFEF}95.4$\pm$0.9  \\
Capsule     & 70.2$\pm$3.0  & 57.7$\pm$7.3 & 67.8$\pm$2.9  & 72.3$\pm$6.8 & \cellcolor[HTML]{EFEFEF}84.6$\pm$6.7  & 70.9$\pm$6.1  & 64.3$\pm$3.0 & 72.8$\pm$7.0  & 77.3$\pm$8.8 & \cellcolor[HTML]{EFEFEF}91.3$\pm$7.7 & 78.9$\pm$5.5  & 65.2$\pm$2.5 & 72.8$\pm$7.0  & 82.3$\pm$8.9  & \cellcolor[HTML]{EFEFEF}91.5$\pm$1.5 \\
Carpet      & 98.1$\pm$0.2  & 96.6$\pm$1.0 & 95.3$\pm$0.8  & 99.8$\pm$0.3 & \cellcolor[HTML]{EFEFEF}100.0$\pm$0.0 & 98.3$\pm$0.4  & 97.8$\pm$0.5 & 96.6$\pm$0.5  & 99.8$\pm$0.3 & \cellcolor[HTML]{EFEFEF}100.0$\pm$0.0 & 98.6$\pm$0.2  & 97.9$\pm$0.4 & 96.6$\pm$0.5  & 100.0$\pm$0.0 & \cellcolor[HTML]{EFEFEF}100.0$\pm$0.0 \\
Grid        & 40.0$\pm$6.8  & 54.2$\pm$6.7 & 63.6$\pm$10.3 & 99.5$\pm$0.3 & \cellcolor[HTML]{EFEFEF}99.8$\pm$0.9  & 41.3$\pm$3.6  & 67.2$\pm$4.2 & 67.7$\pm$8.3  & 99.4$\pm$0.2 & \cellcolor[HTML]{EFEFEF}99.9$\pm$1.2  & 44.6$\pm$6.6  & 68.1$\pm$3.8 & 67.7$\pm$8.3  & 99.6$\pm$0.1  & \cellcolor[HTML]{EFEFEF}98.8$\pm$0.5  \\
Hazelnut    & 95.8$\pm$1.3  & 88.3$\pm$2.6 & 88.3$\pm$2.7  & 97.5$\pm$1.4 & \cellcolor[HTML]{EFEFEF}99.8$\pm$0.8  & 96.2$\pm$2.1  & 90.8$\pm$0.8 & 93.2$\pm$3.8  & 98.3$\pm$0.7 & \cellcolor[HTML]{EFEFEF}100.0$\pm$0.6  & 98.4$\pm$1.3  & 91.9$\pm$1.2 & 93.2$\pm$3.8  & 98.4$\pm$0.4  & \cellcolor[HTML]{EFEFEF}99.8$\pm$0.2  \\
Leather     & 100.0$\pm$0.0 & 97.5$\pm$0.7 & 97.3$\pm$0.7  & 99.9$\pm$0.0 & \cellcolor[HTML]{EFEFEF}100.0$\pm$0.0 & 100.0$\pm$0.0 & 97.5$\pm$0.9 & 97.9$\pm$0.7  & 99.9$\pm$0.0 & \cellcolor[HTML]{EFEFEF}100.0$\pm$0.1  & 100.0$\pm$0.0 & 98.5$\pm$0.2 & 97.9$\pm$0.7  & 100.0$\pm$0.0 & \cellcolor[HTML]{EFEFEF}100.0$\pm$0.1  \\
Metal nut   & 71.0$\pm$2.2  & 53.0$\pm$3.8 & 73.4$\pm$2.9  & 98.7$\pm$0.8 & \cellcolor[HTML]{EFEFEF}99.1$\pm$0.5  & 77.0$\pm$7.9  & 54.8$\pm$3.8 & 77.7$\pm$8.5  & 99.4$\pm$0.2 & \cellcolor[HTML]{EFEFEF}98.6$\pm$0.7  & 77.8$\pm$5.7  & 60.7$\pm$5.2 & 77.7$\pm$8.5  & 99.5$\pm$0.2  & \cellcolor[HTML]{EFEFEF}100.0$\pm$0.1  \\
Pill        & 86.5$\pm$3.1  & 61.3$\pm$3.8 & 81.9$\pm$2.8  & 91.2$\pm$2.1 & \cellcolor[HTML]{EFEFEF}92.6$\pm$1.5  & 84.8$\pm$0.9  & 59.1$\pm$6.4 & 82.9$\pm$2.9  & 92.3$\pm$0.7 & \cellcolor[HTML]{EFEFEF}93.6$\pm$0.8 & 86.7$\pm$0.3  & 54.9$\pm$2.7 & 82.9$\pm$2.9  & 92.8$\pm$1.0  & \cellcolor[HTML]{EFEFEF}92.9$\pm$0.8  \\
Screw       & 46.7$\pm$2.5  & 55.0$\pm$2.5 & 44.4$\pm$4.6  & 86.4$\pm$0.9 & \cellcolor[HTML]{EFEFEF}65.0$\pm$2.9  & 46.6$\pm$2.2  & 54.0$\pm$4.4 & 49.0$\pm$3.8  & 86.0$\pm$2.1 & \cellcolor[HTML]{EFEFEF}71.0$\pm$2.0  & 50.5$\pm$5.4  & 50.0$\pm$4.1 & 49.0$\pm$3.8  & 87.9$\pm$1.2  & \cellcolor[HTML]{EFEFEF}83.6$\pm$2.3 \\
Tile        & 99.9$\pm$0.1  & 92.2$\pm$2.2 & 99.0$\pm$0.9  & 99.9$\pm$0.0 & \cellcolor[HTML]{EFEFEF}100.0$\pm$0.0 & 99.9$\pm$0.1  & 93.3$\pm$1.1 & 98.5$\pm$1.0  & 99.9$\pm$0.2 & \cellcolor[HTML]{EFEFEF}100.0$\pm$0.1  & 100.0$\pm$0.0 & 93.1$\pm$0.6 & 98.5$\pm$1.0  & 99.9$\pm$0.1  & \cellcolor[HTML]{EFEFEF}100.0$\pm$0.1  \\
Toothbrush  & 71.7$\pm$2.6  & 82.5$\pm$1.2 & 83.3$\pm$3.8  & 92.2$\pm$4.9 & \cellcolor[HTML]{EFEFEF}98.9$\pm$1.0  & 78.6$\pm$3.2  & 87.6$\pm$4.2 & 85.9$\pm$3.5  & 97.5$\pm$1.6 & \cellcolor[HTML]{EFEFEF}97.5$\pm$1.5  & 78.8$\pm$5.2  & 89.2$\pm$2.5 & 85.9$\pm$3.5  & 96.7$\pm$2.6  & \cellcolor[HTML]{EFEFEF}98.1$\pm$0.6  \\
Transistor  & 77.2$\pm$2.0  & 73.3$\pm$6.0 & 78.1$\pm$6.9  & 83.4$\pm$3.8 & \cellcolor[HTML]{EFEFEF}94.0$\pm$6.5  & 81.3$\pm$3.7  & 72.8$\pm$6.3 & 90.0$\pm$4.3  & 85.3$\pm$1.7 & \cellcolor[HTML]{EFEFEF}97.4$\pm$4.0  & 81.4$\pm$2.1  & 82.4$\pm$6.5 & 90.0$\pm$4.3  & 85.7$\pm$2.5  & \cellcolor[HTML]{EFEFEF}95.6$\pm$3.8  \\
Wood        & 98.8$\pm$0.3  & 96.1$\pm$1.2 & 97.8$\pm$0.3  & 99.9$\pm$0.1 & \cellcolor[HTML]{EFEFEF}97.9$\pm$0.3  & 99.2$\pm$0.4  & 96.9$\pm$0.5 & 98.3$\pm$0.6  & 99.9$\pm$0.1 & \cellcolor[HTML]{EFEFEF}98.7$\pm$0.1  & 98.9$\pm$0.6  & 97.0$\pm$0.2 & 98.3$\pm$0.6  & 99.8$\pm$0.3  & \cellcolor[HTML]{EFEFEF}98.6$\pm$0.3  \\
Zipper      & 89.3$\pm$1.9  & 85.8$\pm$2.7 & 92.3$\pm$0.5  & 88.8$\pm$5.9 & \cellcolor[HTML]{EFEFEF}93.9$\pm$3.2  & 93.3$\pm$2.9  & 86.3$\pm$2.6 & 94.0$\pm$2.1  & 94.0$\pm$1.4 & \cellcolor[HTML]{EFEFEF}95.8$\pm$1.7  & 95.1$\pm$1.3  & 88.3$\pm$2.0 & 94.0$\pm$2.1  & 94.5$\pm$0.5  & \cellcolor[HTML]{EFEFEF}95.0$\pm$2.3  \\ \midrule
Mean        & 81.0$\pm$2.0  & 76.6$\pm$3.1 & 83.4$\pm$3.0  & \underline{93.1$\pm$2.0} & \cellcolor[HTML]{EFEFEF}\textbf{94.6$\pm$1.7}  & 82.9$\pm$2.6  & 78.9$\pm$3.1 & 86.3$\pm$3.3  & \underline{94.4$\pm$1.3} & \cellcolor[HTML]{EFEFEF}\textbf{95.7$\pm$1.5}  & 84.8$\pm$2.5  & 80.4$\pm$2.5 & 88.8$\pm$2.6  & \underline{95.2$\pm$1.3 } & \cellcolor[HTML]{EFEFEF}\textbf{96.6$\pm$0.9}  \\ \bottomrule
\end{tabular}
}
\caption{Comparison of image-level anomaly detection in terms of subset-wise AUROC on MVTec.}
\label{MVTec-I-ROC}
\end{table*}

\begin{table*}[]
\centering
\resizebox{\linewidth}{!}{
\begin{tabular}{c|ccccc|ccccc|ccccc}
\toprule
MVTec & \multicolumn{5}{c|}{1-shot} & \multicolumn{5}{c|}{2-shot} & \multicolumn{5}{c}{4-shot} \\
Pixel-AUROC & SPADE    & PaDiM    & PatchCore & WinCLIP+  & \cellcolor[HTML]{EFEFEF}PromptAD & SPADE    & PaDiM    & PatchCore & WinCLIP+  & \cellcolor[HTML]{EFEFEF}PromptAD & SPADE    & PaDiM    & PatchCore & WinCLIP+  & \cellcolor[HTML]{EFEFEF}PromptAD \\ \midrule
Bottle      & 95.3$\pm$0.2 & 96.1$\pm$0.5 & 97.9$\pm$0.1  & 97.5$\pm$0.2 & \cellcolor[HTML]{EFEFEF}99.6$\pm$0.2 & 95.7$\pm$0.2 & 96.9$\pm$0.1 & 98.1$\pm$0.0  & 97.7$\pm$0.1 & \cellcolor[HTML]{EFEFEF}99.5$\pm$0.1 & 96.1$\pm$0.0 & 97.1$\pm$0.1 & 98.2$\pm$0.0  & 97.8$\pm$0.0 & \cellcolor[HTML]{EFEFEF}99.5$\pm$0.1 \\
Cable       & 86.4$\pm$0.2 & 88.4$\pm$1.2 & 95.5$\pm$0.8  & 93.8$\pm$0.6 & \cellcolor[HTML]{EFEFEF}98.4$\pm$0.5 & 87.4$\pm$0.3 & 90.0$\pm$0.8 & 96.4$\pm$0.3  & 94.3$\pm$0.4 & \cellcolor[HTML]{EFEFEF}97.6$\pm$0.3 & 88.2$\pm$0.2 & 92.1$\pm$0.4 & 97.5$\pm$0.3  & 94.9$\pm$0.1 & \cellcolor[HTML]{EFEFEF}98.2$\pm$0.2 \\
Capsule     & 96.3$\pm$0.2 & 94.5$\pm$0.6 & 95.6$\pm$0.4  & 94.6$\pm$0.8 & \cellcolor[HTML]{EFEFEF}99.5$\pm$0.7 & 96.7$\pm$0.1 & 95.2$\pm$0.5 & 96.5$\pm$0.4  & 96.4$\pm$0.3 & \cellcolor[HTML]{EFEFEF}99.3$\pm$0.5 & 97.0$\pm$0.2 & 96.2$\pm$0.4 & 96.8$\pm$0.6  & 96.2$\pm$0.5 & \cellcolor[HTML]{EFEFEF}99.3$\pm$0.3 \\
Carpet      & 98.2$\pm$0.0 & 97.8$\pm$0.2 & 98.4$\pm$0.1  & 99.4$\pm$0.0 & \cellcolor[HTML]{EFEFEF}95.9$\pm$0.0 & 98.3$\pm$0.0 & 98.2$\pm$0.0 & 98.5$\pm$0.1  & 99.3$\pm$0.0 & \cellcolor[HTML]{EFEFEF}96.1$\pm$0.0 & 98.4$\pm$0.0 & 98.4$\pm$0.0 & 98.6$\pm$0.1  & 99.3$\pm$0.0 & \cellcolor[HTML]{EFEFEF}96.2$\pm$0.0 \\
Grid        & 80.7$\pm$1.3 & 70.2$\pm$2.8 & 58.8$\pm$4.9  & 96.8$\pm$1.0 & \cellcolor[HTML]{EFEFEF}95.1$\pm$0.5 & 83.5$\pm$1.0 & 70.8$\pm$2.0 & 62.6$\pm$3.2  & 97.7$\pm$0.8 & \cellcolor[HTML]{EFEFEF}95.5$\pm$0.4 & 87.2$\pm$1.1 & 77.0$\pm$1.8 & 69.4$\pm$1.3  & 98.0$\pm$0.2 & \cellcolor[HTML]{EFEFEF}95.2$\pm$0.3 \\
Hazelnut    & 97.2$\pm$0.1 & 95.4$\pm$0.7 & 95.8$\pm$0.6  & 98.5$\pm$0.2 & \cellcolor[HTML]{EFEFEF}97.3$\pm$0.4 & 97.6$\pm$0.1 & 96.8$\pm$0.3 & 96.3$\pm$0.6  & 98.7$\pm$0.1 & \cellcolor[HTML]{EFEFEF}97.6$\pm$0.3 & 97.7$\pm$0.1 & 97.2$\pm$0.2 & 97.6$\pm$0.1  & 98.8$\pm$0.0 & \cellcolor[HTML]{EFEFEF}97.9$\pm$0.2 \\
Leather     & 99.1$\pm$0.0 & 98.5$\pm$0.1 & 98.8$\pm$0.2  & 99.3$\pm$0.0 & \cellcolor[HTML]{EFEFEF}93.3$\pm$0.0 & 99.1$\pm$0.0 & 98.7$\pm$0.1 & 99.0$\pm$0.1  & 99.3$\pm$0.0 & \cellcolor[HTML]{EFEFEF}93.2$\pm$0.0 & 99.1$\pm$0.0 & 98.8$\pm$0.0 & 99.1$\pm$0.0  & 99.3$\pm$0.0 & \cellcolor[HTML]{EFEFEF}93.9$\pm$0.0 \\
Metal nut   & 83.8$\pm$0.7 & 74.6$\pm$1.1 & 89.3$\pm$1.4  & 90.0$\pm$0.6 & \cellcolor[HTML]{EFEFEF}97.3$\pm$0.7 & 85.8$\pm$1.1 & 80.3$\pm$2.1 & 94.6$\pm$1.4  & 91.4$\pm$0.4 & \cellcolor[HTML]{EFEFEF}97.4$\pm$0.7 & 87.1$\pm$0.7 & 82.7$\pm$3.9 & 95.9$\pm$1.8  & 92.9$\pm$0.4 & \cellcolor[HTML]{EFEFEF}97.7$\pm$0.5 \\
Pill        & 89.4$\pm$0.4 & 84.8$\pm$1.0 & 93.1$\pm$1.1  & 96.4$\pm$0.3 & \cellcolor[HTML]{EFEFEF}98.4$\pm$0.4 & 89.9$\pm$0.2 & 87.3$\pm$0.7 & 94.2$\pm$0.3  & 97.0$\pm$0.2 & \cellcolor[HTML]{EFEFEF}98.5$\pm$0.3 & 90.7$\pm$0.2 & 88.9$\pm$0.5 & 94.8$\pm$0.4  & 97.1$\pm$0.0 & \cellcolor[HTML]{EFEFEF}98.5$\pm$0.2 \\
Screw       & 94.8$\pm$0.2 & 83.3$\pm$0.7 & 89.6$\pm$0.5  & 94.5$\pm$0.4 & \cellcolor[HTML]{EFEFEF}91.4$\pm$0.5 & 95.6$\pm$0.4 & 89.8$\pm$0.8 & 90.0$\pm$0.7  & 95.2$\pm$0.3 & \cellcolor[HTML]{EFEFEF}95.1$\pm$0.7 & 96.4$\pm$0.4 & 90.8$\pm$0.2 & 91.3$\pm$1.0  & 96.0$\pm$0.5 & \cellcolor[HTML]{EFEFEF}94.9$\pm$0.8 \\
Tile        & 91.7$\pm$0.3 & 84.1$\pm$1.1 & 94.1$\pm$0.5  & 96.3$\pm$0.2 & \cellcolor[HTML]{EFEFEF}92.8$\pm$0.1 & 92.0$\pm$0.1 & 87.7$\pm$0.2 & 94.4$\pm$0.2  & 96.5$\pm$0.1 & \cellcolor[HTML]{EFEFEF}94.1$\pm$0.1 & 92.2$\pm$0.1 & 88.9$\pm$0.3 & 94.6$\pm$0.1  & 96.6$\pm$0.1 & \cellcolor[HTML]{EFEFEF}94.1$\pm$0.1 \\
Toothbrush  & 94.6$\pm$0.6 & 97.3$\pm$0.3 & 97.3$\pm$0.4  & 97.8$\pm$0.1 & \cellcolor[HTML]{EFEFEF}94.0$\pm$0.2 & 96.2$\pm$0.3 & 97.7$\pm$0.3 & 97.5$\pm$0.2  & 98.1$\pm$0.1 & \cellcolor[HTML]{EFEFEF}95.5$\pm$0.1 & 97.0$\pm$0.6 & 98.4$\pm$0.2 & 98.4$\pm$0.4  & 98.4$\pm$0.5 & \cellcolor[HTML]{EFEFEF}96.2$\pm$0.0 \\
Transistor  & 71.4$\pm$1.3 & 90.2$\pm$2.8 & 84.9$\pm$2.7  & 85.0$\pm$1.8 & \cellcolor[HTML]{EFEFEF}99.1$\pm$1.9 & 72.8$\pm$0.9 & 92.3$\pm$2.1 & 89.6$\pm$0.9  & 88.3$\pm$1.0 & \cellcolor[HTML]{EFEFEF}99.0$\pm$1.1 & 73.4$\pm$0.7 & 94.0$\pm$2.7 & 90.7$\pm$1.4  & 88.5$\pm$1.2 & \cellcolor[HTML]{EFEFEF}99.0$\pm$0.6 \\
Wood        & 93.4$\pm$0.1 & 90.7$\pm$0.4 & 92.7$\pm$0.9  & 94.6$\pm$1.0 & \cellcolor[HTML]{EFEFEF}89.4$\pm$0.3 & 93.8$\pm$0.1 & 91.9$\pm$0.1 & 93.2$\pm$0.7  & 95.3$\pm$0.4 & \cellcolor[HTML]{EFEFEF}89.1$\pm$0.2 & 93.9$\pm$0.1 & 92.2$\pm$0.1 & 93.5$\pm$0.3  & 95.4$\pm$0.2 & \cellcolor[HTML]{EFEFEF}90.6$\pm$0.1 \\
Zipper      & 94.9$\pm$0.3 & 93.9$\pm$0.8 & 97.4$\pm$0.4  & 93.9$\pm$0.8 & \cellcolor[HTML]{EFEFEF}96.6$\pm$0.5 & 95.8$\pm$0.2 & 95.4$\pm$0.3 & 98.0$\pm$0.1  & 94.1$\pm$0.7 & \cellcolor[HTML]{EFEFEF}95.5$\pm$0.4 & 96.2$\pm$0.1 & 96.1$\pm$0.2 & 98.1$\pm$0.1  & 94.2$\pm$0.4 & \cellcolor[HTML]{EFEFEF}96.6$\pm$0.3 \\ \midrule
Mean        & 91.2$\pm$0.4 & 89.3$\pm$0.9 & 92.0$\pm$1.0  & \underline{95.2$\pm$0.5} & \cellcolor[HTML]{EFEFEF}\textbf{95.9$\pm$0.5} & 92.0$\pm$0.3 & 91.3$\pm$0.7 & 93.3$\pm$0.6  & \underline{96.0$\pm$0.3} & \cellcolor[HTML]{EFEFEF}\textbf{96.2$\pm$0.3} & 92.7$\pm$0.3 & 92.6$\pm$0.7 & 94.3$\pm$0.5  & \underline{96.2$\pm$0.3} & \cellcolor[HTML]{EFEFEF}\textbf{96.5$\pm$0.2} \\ \bottomrule
\end{tabular}
}
\caption{Comparison of pixel-level anomaly detection in terms of subset-wise AUROC on MVTec.}
\label{MVTec-P-ROC}
\end{table*}

\begin{table*}[]
\centering
\resizebox{\linewidth}{!}{
\begin{tabular}{c|ccccc|ccccc|ccccc}
\toprule
VisA & \multicolumn{5}{c|}{1-shot} & \multicolumn{5}{c|}{2-shot} & \multicolumn{5}{c}{4-shot} \\
Image-AUROC & SPADE     & PaDiM     & PatchCore & WinCLIP+   & \cellcolor[HTML]{EFEFEF}PromptAD & SPADE     & PaDiM     & PatchCore & WinCLIP+  & \cellcolor[HTML]{EFEFEF}PromptAD & SPADE    & PaDiM    & PatchCore & WinCLIP+  & \cellcolor[HTML]{EFEFEF}PromptAD \\ \midrule
Candle      & 86.1$\pm$5.6  & 70.8$\pm$4.1  & 85.1$\pm$1.4  & 93.4$\pm$1.4  & \cellcolor[HTML]{EFEFEF}90.3$\pm$2.7 & 91.3$\pm$3.3  & 75.8$\pm$2.1  & 85.3$\pm$1.5  & 94.8$\pm$1.0 & \cellcolor[HTML]{EFEFEF}91.0$\pm$1.2 & 92.8$\pm$2.1 & 77.5$\pm$1.6 & 87.8$\pm$0.8  & 95.1$\pm$0.3 & \cellcolor[HTML]{EFEFEF}93.0$\pm$1.2 \\
Capsules    & 73.3$\pm$7.5  & 51.0$\pm$7.8  & 60.0$\pm$7.6  & 85.0$\pm$3.1  & \cellcolor[HTML]{EFEFEF}84.5$\pm$3.5 & 71.7$\pm$11.2 & 51.7$\pm$4.6  & 57.8$\pm$5.4  & 84.9$\pm$0.8 & \cellcolor[HTML]{EFEFEF}84.9$\pm$3.7 & 73.4$\pm$7.1 & 52.7$\pm$3.4 & 63.4$\pm$5.4  & 86.8$\pm$1.7 & \cellcolor[HTML]{EFEFEF}80.6$\pm$2.1 \\
Cashew      & 95.9$\pm$1.1  & 62.3$\pm$9.9  & 89.5$\pm$4.4  & 94.0$\pm$0.4  & \cellcolor[HTML]{EFEFEF}95.6$\pm$1.0 & 97.3$\pm$1.4  & 74.6$\pm$3.6  & 93.6$\pm$0.6  & 94.3$\pm$0.5 & \cellcolor[HTML]{EFEFEF}94.7$\pm$1.4 & 96.4$\pm$1.3 & 77.7$\pm$3.2 & 93.0$\pm$1.5  & 95.2$\pm$0.8 & \cellcolor[HTML]{EFEFEF}93.6$\pm$2.2 \\
Chewinggum  & 92.1$\pm$2.0  & 69.9$\pm$4.9  & 97.3$\pm$0.3  & 97.6$\pm$0.8  & \cellcolor[HTML]{EFEFEF}96.4$\pm$1.2 & 93.4$\pm$1.0  & 82.7$\pm$2.1  & 97.8$\pm$0.6  & 97.3$\pm$0.8 & \cellcolor[HTML]{EFEFEF}96.6$\pm$0.6 & 93.5$\pm$1.4 & 83.5$\pm$3.7 & 98.3$\pm$0.3  & 97.7$\pm$0.3 & \cellcolor[HTML]{EFEFEF}96.8$\pm$0.4 \\
Fryum       & 81.1$\pm$4.0  & 58.3$\pm$5.9  & 75.0$\pm$4.8  & 88.5$\pm$1.9  & \cellcolor[HTML]{EFEFEF}90.3$\pm$1.8 & 90.5$\pm$3.9  & 69.2$\pm$9.0  & 83.4$\pm$2.4  & 90.5$\pm$0.4 & \cellcolor[HTML]{EFEFEF}89.2$\pm$0.9 & 92.9$\pm$1.6 & 71.2$\pm$5.9 & 88.6$\pm$1.3  & 90.8$\pm$0.5 & \cellcolor[HTML]{EFEFEF}89.0$\pm$2.3 \\
Macaroni1   & 66.0$\pm$10.5 & 62.1$\pm$4.6  & 68.0$\pm$3.4  & 82.9$\pm$1.5  & \cellcolor[HTML]{EFEFEF}88.6$\pm$3.1 & 69.1$\pm$8.2  & 62.2$\pm$5.0  & 75.6$\pm$4.6  & 83.3$\pm$1.9 & \cellcolor[HTML]{EFEFEF}84.2$\pm$2.5 & 65.8$\pm$1.2 & 65.9$\pm$3.9 & 82.9$\pm$2.7  & 85.2$\pm$0.9 & \cellcolor[HTML]{EFEFEF}88.2$\pm$2.5 \\
Macaroni2   & 55.8$\pm$6.1  & 47.5$\pm$5.9  & 55.6$\pm$4.6  & 70.2$\pm$0.9  & \cellcolor[HTML]{EFEFEF}69.1$\pm$3.0 & 58.3$\pm$4.4  & 50.8$\pm$2.9  & 57.3$\pm$5.6  & 71.8$\pm$2.0 & \cellcolor[HTML]{EFEFEF}82.6$\pm$0.9 & 56.7$\pm$3.2 & 55.0$\pm$2.9 & 61.7$\pm$1.8  & 70.9$\pm$2.2 & \cellcolor[HTML]{EFEFEF}81.2$\pm$1.8 \\
PCB1        & 87.2$\pm$2.3  & 76.2$\pm$1.2  & 78.9$\pm$1.1  & 75.6$\pm$23.0 & \cellcolor[HTML]{EFEFEF}88.7$\pm$0.7 & 86.7$\pm$1.1  & 62.4$\pm$10.8 & 71.5$\pm$20.0 & 76.7$\pm$5.2 & \cellcolor[HTML]{EFEFEF}90.9$\pm$5.4 & 83.4$\pm$8.5 & 82.6$\pm$1.5 & 84.7$\pm$6.7  & 88.3$\pm$1.7 & \cellcolor[HTML]{EFEFEF}90.9$\pm$2.5 \\
PCB2        & 73.5$\pm$3.7  & 61.2$\pm$2.0  & 81.5$\pm$0.8  & 62.2$\pm$3.9  & \cellcolor[HTML]{EFEFEF}71.6$\pm$4.1 & 70.3$\pm$8.1  & 66.8$\pm$2.0  & 84.3$\pm$1.7  & 62.6$\pm$3.7 & \cellcolor[HTML]{EFEFEF}73.0$\pm$3.0 & 71.7$\pm$7.0 & 73.5$\pm$2.4 & 84.3$\pm$1.0  & 67.5$\pm$2.6 & \cellcolor[HTML]{EFEFEF}78.6$\pm$2.5 \\
PCB3        & 72.2$\pm$1.0  & 51.4$\pm$12.2 & 82.7$\pm$2.3  & 74.1$\pm$1.1  & \cellcolor[HTML]{EFEFEF}79.1$\pm$3.6 & 75.8$\pm$5.7  & 67.3$\pm$3.8  & 84.8$\pm$1.2  & 78.8$\pm$1.9 & \cellcolor[HTML]{EFEFEF}76.2$\pm$2.2 & 79.0$\pm$4.1 & 65.9$\pm$1.9 & 87.0$\pm$1.1  & 83.3$\pm$1.7 & \cellcolor[HTML]{EFEFEF}80.3$\pm$1.7 \\
PCB4        & 93.4$\pm$1.3  & 76.1$\pm$3.6  & 93.9$\pm$2.8  & 85.2$\pm$8.9  & \cellcolor[HTML]{EFEFEF}91.4$\pm$3.2 & 86.1$\pm$8.2  & 69.3$\pm$13.7 & 94.3$\pm$3.2  & 82.3$\pm$9.9 & \cellcolor[HTML]{EFEFEF}97.5$\pm$2.4 & 95.4$\pm$2.3 & 85.4$\pm$2.0 & 95.6$\pm$1.6  & 87.6$\pm$8.0 & \cellcolor[HTML]{EFEFEF}97.8$\pm$1.4 \\
Pipe fryum  & 77.9$\pm$3.2  & 66.7$\pm$2.2  & 90.7$\pm$1.7  & 97.2$\pm$1.1  & \cellcolor[HTML]{EFEFEF}96.9$\pm$0.2 & 78.1$\pm$3.0  & 75.3$\pm$1.8  & 93.5$\pm$1.3  & 98.0$\pm$0.6 & \cellcolor[HTML]{EFEFEF}98.9$\pm$0.3 & 79.3$\pm$0.9 & 82.9$\pm$2.2 & 96.4$\pm$0.7  & 98.5$\pm$0.4 & \cellcolor[HTML]{EFEFEF}98.6$\pm$0.2 \\ \midrule
Mean        & 79.5$\pm$4.0  & 62.8$\pm$5.4  & 79.9$\pm$2.9  & \underline{83.8$\pm$4.0}  & \cellcolor[HTML]{EFEFEF}\textbf{86.9$\pm$2.3} & 80.7$\pm$5.0  & 67.4$\pm$5.1  & 81.6$\pm$4.0  & \underline{84.6$\pm$2.4} & \cellcolor[HTML]{EFEFEF}\textbf{88.3$\pm$2.0} & 81.7$\pm$3.4 & 72.8$\pm$2.9 & 85.3$\pm$2.1  & \underline{87.3$\pm$1.8} & \cellcolor[HTML]{EFEFEF}\textbf{89.1$\pm$1.7} \\ \bottomrule
\end{tabular}
}
\caption{Comparison of image-level anomaly detection in terms of subset-wise AUROC on VisA.}
\label{VisA-I-ROC}
\end{table*}

\begin{table*}[]
\centering
\resizebox{\linewidth}{!}{
\begin{tabular}{c|ccccc|ccccc|ccccc}
\toprule
VisA & \multicolumn{5}{c|}{1-shot} & \multicolumn{5}{c|}{2-shot} & \multicolumn{5}{c}{4-shot} \\
Pixel-AUROC & SPADE    & PaDiM    & PatchCore & WinCLIP+  & \cellcolor[HTML]{EFEFEF}PromptAD & SPADE    & PaDiM    & PatchCore & WinCLIP+  & \cellcolor[HTML]{EFEFEF}PromptAD & SPADE    & PaDiM    & PatchCore & WinCLIP+  & \cellcolor[HTML]{EFEFEF}PromptAD \\ \midrule
Candle      & 97.9$\pm$0.3 & 91.7$\pm$2.2 & 97.2$\pm$0.2  & 97.4$\pm$0.2 & \cellcolor[HTML]{EFEFEF}95.8$\pm$0.2 & 98.1$\pm$0.2 & 94.9$\pm$0.8 & 97.7$\pm$0.3  & 97.7$\pm$0.1 & \cellcolor[HTML]{EFEFEF}95.9$\pm$0.1 & 98.2$\pm$0.1 & 95.4$\pm$0.2 & 97.9$\pm$0.1  & 97.8$\pm$0.2 & \cellcolor[HTML]{EFEFEF}96.0$\pm$0.1 \\
Capsules    & 95.5$\pm$0.5 & 70.9$\pm$1.1 & 93.2$\pm$0.9  & 96.4$\pm$0.6 & \cellcolor[HTML]{EFEFEF}95.4$\pm$0.9 & 96.5$\pm$0.9 & 75.7$\pm$1.7 & 94.0$\pm$0.2  & 96.8$\pm$0.3 & \cellcolor[HTML]{EFEFEF}96.1$\pm$0.7 & 97.7$\pm$0.1 & 79.1$\pm$0.7 & 94.8$\pm$0.5  & 97.1$\pm$0.2 & \cellcolor[HTML]{EFEFEF}96.8$\pm$0.6 \\
Cashew      & 95.9$\pm$0.5 & 95.5$\pm$0.6 & 98.1$\pm$0.1  & 98.5$\pm$0.2 & \cellcolor[HTML]{EFEFEF}99.1$\pm$0.2 & 95.9$\pm$0.4 & 96.4$\pm$0.4 & 98.2$\pm$0.2  & 98.5$\pm$0.1 & \cellcolor[HTML]{EFEFEF}99.2$\pm$0.1 & 95.9$\pm$0.3 & 97.2$\pm$0.3 & 98.3$\pm$0.2  & 98.7$\pm$0.0 & \cellcolor[HTML]{EFEFEF}99.2$\pm$0.1 \\
Chewinggum  & 96.0$\pm$0.4 & 90.1$\pm$0.4 & 96.9$\pm$0.3  & 98.6$\pm$0.1 & \cellcolor[HTML]{EFEFEF}99.1$\pm$0.1 & 96.0$\pm$0.3 & 93.1$\pm$0.7 & 96.6$\pm$0.1  & 98.6$\pm$0.1 & \cellcolor[HTML]{EFEFEF}99.2$\pm$0.1 & 95.7$\pm$0.3 & 94.4$\pm$0.5 & 96.8$\pm$0.1  & 98.5$\pm$0.1 & \cellcolor[HTML]{EFEFEF}99.2$\pm$0.2 \\
Fryum       & 93.5$\pm$0.3 & 93.3$\pm$0.6 & 93.3$\pm$0.5  & 96.4$\pm$0.3 & \cellcolor[HTML]{EFEFEF}95.4$\pm$0.3 & 93.9$\pm$0.2 & 94.1$\pm$0.6 & 94.0$\pm$0.3  & 97.0$\pm$0.2 & \cellcolor[HTML]{EFEFEF}96.4$\pm$0.2 & 94.4$\pm$0.1 & 95.0$\pm$0.4 & 94.2$\pm$0.2  & 97.1$\pm$0.1 & \cellcolor[HTML]{EFEFEF}96.6$\pm$0.2 \\
Macaroni1   & 97.9$\pm$0.2 & 89.4$\pm$0.9 & 95.2$\pm$0.4  & 96.4$\pm$0.6 & \cellcolor[HTML]{EFEFEF}97.8$\pm$0.1 & 98.5$\pm$0.2 & 91.7$\pm$0.3 & 96.0$\pm$1.3  & 96.5$\pm$0.7 & \cellcolor[HTML]{EFEFEF}98.3$\pm$0.1 & 98.8$\pm$0.1 & 93.5$\pm$0.5 & 97.0$\pm$0.3  & 97.0$\pm$0.2 & \cellcolor[HTML]{EFEFEF}98.2$\pm$0.1 \\
Macaroni2   & 94.1$\pm$1.0 & 86.4$\pm$1.1 & 89.1$\pm$1.6  & 96.8$\pm$0.4 & \cellcolor[HTML]{EFEFEF}96.6$\pm$0.5 & 95.2$\pm$0.4 & 90.1$\pm$0.8 & 90.2$\pm$1.9  & 96.8$\pm$0.6 & \cellcolor[HTML]{EFEFEF}97.2$\pm$0.3 & 96.4$\pm$0.2 & 90.2$\pm$0.3 & 93.9$\pm$0.3  & 97.3$\pm$0.3 & \cellcolor[HTML]{EFEFEF}97.0$\pm$0.3 \\
PCB1        & 94.7$\pm$0.4 & 89.9$\pm$0.3 & 96.1$\pm$1.5  & 96.6$\pm$0.6 & \cellcolor[HTML]{EFEFEF}96.6$\pm$0.8 & 96.5$\pm$1.5 & 90.6$\pm$0.6 & 97.6$\pm$0.9  & 97.0$\pm$0.9 & \cellcolor[HTML]{EFEFEF}96.9$\pm$0.4 & 96.8$\pm$1.5 & 93.2$\pm$1.5 & 98.1$\pm$1.0  & 98.1$\pm$0.9 & \cellcolor[HTML]{EFEFEF}98.2$\pm$0.3 \\
PCB2        & 95.1$\pm$0.2 & 90.9$\pm$1.4 & 95.4$\pm$0.2  & 93.0$\pm$0.4 & \cellcolor[HTML]{EFEFEF}93.5$\pm$0.9 & 95.7$\pm$0.1 & 93.9$\pm$0.9 & 96.0$\pm$0.3  & 93.9$\pm$0.2 & \cellcolor[HTML]{EFEFEF}94.8$\pm$0.8 & 96.3$\pm$0.0 & 93.7$\pm$1.0 & 96.6$\pm$0.2  & 94.6$\pm$0.4 & \cellcolor[HTML]{EFEFEF}95.3$\pm$0.5 \\
PCB3        & 96.0$\pm$0.1 & 93.9$\pm$0.3 & 96.2$\pm$0.3  & 94.3$\pm$0.3 & \cellcolor[HTML]{EFEFEF}95.9$\pm$0.5 & 96.6$\pm$0.1 & 95.1$\pm$0.5 & 97.1$\pm$0.1  & 95.1$\pm$0.2 & \cellcolor[HTML]{EFEFEF}96.1$\pm$0.4 & 96.9$\pm$0.0 & 95.7$\pm$0.1 & 97.4$\pm$0.2  & 95.8$\pm$0.1 & \cellcolor[HTML]{EFEFEF}96.8$\pm$0.2 \\
PCB4        & 92.0$\pm$0.6 & 89.6$\pm$0.6 & 95.6$\pm$0.6  & 94.0$\pm$0.9 & \cellcolor[HTML]{EFEFEF}95.5$\pm$0.5 & 92.8$\pm$0.3 & 90.7$\pm$0.9 & 96.2$\pm$0.4  & 95.6$\pm$0.3 & \cellcolor[HTML]{EFEFEF}95.6$\pm$0.3 & 94.1$\pm$0.2 & 92.1$\pm$0.5 & 97.0$\pm$0.2  & 96.1$\pm$0.3 & \cellcolor[HTML]{EFEFEF}96.2$\pm$0.4 \\
Pipe fryum  & 98.4$\pm$0.2 & 97.2$\pm$0.6 & 98.8$\pm$0.2  & 98.3$\pm$0.2 & \cellcolor[HTML]{EFEFEF}99.1$\pm$0.2 & 98.7$\pm$0.1 & 98.1$\pm$0.4 & 99.1$\pm$0.1  & 98.5$\pm$0.2 & \cellcolor[HTML]{EFEFEF}99.4$\pm$0.2 & 98.8$\pm$0.0 & 98.5$\pm$0.1 & 99.1$\pm$0.0  & 98.7$\pm$0.1 & \cellcolor[HTML]{EFEFEF}99.3$\pm$0.3 \\ \midrule
Mean        & 95.6$\pm$0.4 & 89.9$\pm$0.8 & 95.4$\pm$0.6  & \underline{96.4$\pm$0.4} & \cellcolor[HTML]{EFEFEF}\textbf{96.7$\pm$0.4} & 96.2$\pm$0.4 & 92.0$\pm$0.7 & 96.1$\pm$0.5  & \underline{96.8$\pm$0.3} & \cellcolor[HTML]{EFEFEF}\textbf{97.1$\pm$0.3} & 96.6$\pm$0.3 & 93.2$\pm$0.5 & 96.8$\pm$0.3  & \underline{97.2$\pm$0.2} & \cellcolor[HTML]{EFEFEF}\textbf{97.4$\pm$0.3} \\ \bottomrule
\end{tabular}
}
\caption{Comparison of pixel-level anomaly detection in terms of subset-wise AUROC on VisA.}
\label{VisA-P-ROC}
\end{table*}

\begin{table*}[]
\centering
\resizebox{\linewidth}{!}{
\begin{tabular}{c|ccccc|ccccc|ccccc}
\toprule
MVTec & \multicolumn{5}{c|}{1-shot} & \multicolumn{5}{c|}{2-shot} & \multicolumn{5}{c}{4-shot} \\
Image-level AUPR & SPADE     & PaDiM    & PatchCore & WinCLIP+   & \cellcolor[HTML]{EFEFEF}PromptAD          & SPADE     & PaDiM    & PatchCore & WinCLIP+   & \cellcolor[HTML]{EFEFEF}PromptAD          & SPADE     & PaDiM    & PatchCore & WinCLIP+   & \cellcolor[HTML]{EFEFEF}PromptAD          \\ \midrule
Bottle           & 99.6$\pm$0.1  & 99.2$\pm$0.2 & 99.8$\pm$0.1  & 99.4$\pm$0.3  & \cellcolor[HTML]{EFEFEF}99.8$\pm$0.1          & 99.8$\pm$0.0  & 99.6$\pm$0.3 & 99.8$\pm$0.1  & 99.8$\pm$0.1  & \cellcolor[HTML]{EFEFEF}99.9$\pm$0.1          & 99.9$\pm$0.0  & 99.7$\pm$0.0 & 99.8$\pm$0.1  & 99.8$\pm$0.1  & \cellcolor[HTML]{EFEFEF}100.0$\pm$0.1         \\
Cable            & 79.6$\pm$2.3  & 64.9$\pm$3.8 & 93.8$\pm$2.2  & 93.2$\pm$1.1  & \cellcolor[HTML]{EFEFEF}95.5$\pm$0.7          & 84.5$\pm$3.1  & 69.6$\pm$6.6 & 95.1$\pm$1.3  & 92.9$\pm$0.6  & \cellcolor[HTML]{EFEFEF}96.9$\pm$0.7          & 88.8$\pm$1.9  & 76.1$\pm$5.6 & 97.1$\pm$0.7  & 94.4$\pm$0.3  & \cellcolor[HTML]{EFEFEF}97.4$\pm$0.5          \\
Capsule          & 91.2$\pm$0.9  & 86.9$\pm$2.2 & 89.4$\pm$2.0  & 91.6$\pm$2.7  & \cellcolor[HTML]{EFEFEF}97.8$\pm$3.1          & 91.6$\pm$2.1  & 88.4$\pm$0.8 & 91.0$\pm$2.9  & 93.3$\pm$3.6  & \cellcolor[HTML]{EFEFEF}97.0$\pm$3.0          & 94.4$\pm$1.9  & 87.8$\pm$0.8 & 94.9$\pm$1.1  & 95.1$\pm$3.3  & \cellcolor[HTML]{EFEFEF}98.6$\pm$2.2          \\
Carpet           & 99.4$\pm$0.0  & 99.0$\pm$0.2 & 98.7$\pm$0.2  & 99.9$\pm$0.1  & \cellcolor[HTML]{EFEFEF}100.0$\pm$0.0         & 99.5$\pm$0.1  & 99.4$\pm$0.1 & 99.0$\pm$0.1  & 99.9$\pm$0.1  & \cellcolor[HTML]{EFEFEF}100.0$\pm$0.0         & 99.6$\pm$0.1  & 99.4$\pm$0.1 & 98.8$\pm$0.2  & 100.0$\pm$0.0 & \cellcolor[HTML]{EFEFEF}100.0$\pm$0.0         \\
Grid             & 66.9$\pm$2.1  & 75.0$\pm$3.3 & 81.1$\pm$4.9  & 99.9$\pm$0.1  & \cellcolor[HTML]{EFEFEF}98.8$\pm$0.3          & 68.3$\pm$2.1  & 82.5$\pm$2.3 & 84.1$\pm$4.0  & 99.8$\pm$0.1  & \cellcolor[HTML]{EFEFEF}99.9$\pm$0.3          & 68.8$\pm$4.2  & 83.0$\pm$1.8 & 86.4$\pm$4.0  & 99.9$\pm$0.0  & \cellcolor[HTML]{EFEFEF}99.7$\pm$0.1          \\
Hazelnut         & 97.9$\pm$0.6  & 93.3$\pm$1.7 & 92.9$\pm$2.2  & 98.6$\pm$0.7  & \cellcolor[HTML]{EFEFEF}99.7$\pm$0.3          & 98.0$\pm$1.1  & 94.1$\pm$0.5 & 96.0$\pm$2.0  & 99.1$\pm$0.4  & \cellcolor[HTML]{EFEFEF}99.8$\pm$0.2          & 99.1$\pm$0.7  & 94.8$\pm$0.6 & 97.0$\pm$1.2  & 99.1$\pm$0.2  & \cellcolor[HTML]{EFEFEF}99.9$\pm$0.2          \\
Leather          & 100.0$\pm$0.0 & 99.2$\pm$0.2 & 99.1$\pm$0.2  & 100.0$\pm$0.0 & \cellcolor[HTML]{EFEFEF}100.0$\pm$0.0         & 100.0$\pm$0.0 & 99.2$\pm$0.3 & 99.3$\pm$0.2  & 100.0$\pm$0.0 & \cellcolor[HTML]{EFEFEF}100.0$\pm$0.0         & 100.0$\pm$0.0 & 99.6$\pm$0.1 & 99.6$\pm$0.1  & 100.0$\pm$0.0 & \cellcolor[HTML]{EFEFEF}100.0$\pm$0.0         \\
Metal nut        & 91.7$\pm$0.8  & 82.0$\pm$2.7 & 91.0$\pm$1.1  & 99.7$\pm$0.2  & \cellcolor[HTML]{EFEFEF}99.6$\pm$0.1          & 93.7$\pm$2.4  & 82.2$\pm$1.4 & 92.3$\pm$4.0  & 99.9$\pm$0.0  & \cellcolor[HTML]{EFEFEF}100.0$\pm$0.1         & 94.1$\pm$1.8  & 85.5$\pm$1.7 & 97.0$\pm$2.6  & 99.9$\pm$0.1  & \cellcolor[HTML]{EFEFEF}99.9$\pm$0.0          \\
Pill             & 97.0$\pm$0.8  & 88.3$\pm$1.3 & 96.5$\pm$0.6  & 98.3$\pm$0.5  & \cellcolor[HTML]{EFEFEF}98.5$\pm$0.3          & 96.5$\pm$0.4  & 87.9$\pm$2.6 & 96.6$\pm$0.7  & 98.6$\pm$0.1  & \cellcolor[HTML]{EFEFEF}97.8$\pm$0.2          & 97.0$\pm$0.2  & 87.0$\pm$1.2 & 96.9$\pm$0.4  & 98.6$\pm$0.2  & \cellcolor[HTML]{EFEFEF}98.5$\pm$0.1          \\
Screw            & 71.3$\pm$1.8  & 78.1$\pm$1.0 & 71.4$\pm$2.3  & 94.2$\pm$0.6  & \cellcolor[HTML]{EFEFEF}78.5$\pm$3.0          & 71.0$\pm$1.4  & 77.3$\pm$1.3 & 72.9$\pm$3.4  & 94.1$\pm$1.5  & \cellcolor[HTML]{EFEFEF}86.7$\pm$1.9          & 73.7$\pm$2.4  & 75.7$\pm$2.8 & 71.8$\pm$1.9  & 94.9$\pm$0.8  & \cellcolor[HTML]{EFEFEF}93.8$\pm$2.1          \\
Tile             & 100.0$\pm$0.0 & 97.2$\pm$0.7 & 99.6$\pm$0.3  & 100.0$\pm$0.0 & \cellcolor[HTML]{EFEFEF}100.0$\pm$0.0         & 100.0$\pm$0.0 & 97.6$\pm$0.4 & 99.4$\pm$0.4  & 100.0$\pm$0.1 & \cellcolor[HTML]{EFEFEF}100.0$\pm$0.0         & 100.0$\pm$0.0 & 97.6$\pm$0.2 & 99.6$\pm$0.1  & 100.0$\pm$0.0 & \cellcolor[HTML]{EFEFEF}100.0$\pm$0.0         \\
Toothbrush       & 88.3$\pm$0.6  & 93.7$\pm$0.5 & 93.5$\pm$1.4  & 96.7$\pm$2.0  & \cellcolor[HTML]{EFEFEF}98.9$\pm$0.4          & 90.8$\pm$1.3  & 95.2$\pm$1.6 & 94.1$\pm$1.4  & 99.0$\pm$0.6  & \cellcolor[HTML]{EFEFEF}99.3$\pm$0.4          & 91.3$\pm$2.6  & 95.8$\pm$0.7 & 94.8$\pm$0.7  & 98.7$\pm$1.1  & \cellcolor[HTML]{EFEFEF}99.7$\pm$0.1          \\
Transistor       & 76.2$\pm$1.7  & 66.2$\pm$7.5 & 77.7$\pm$5.5  & 79.0$\pm$4.0  & \cellcolor[HTML]{EFEFEF}91.2$\pm$5.9          & 81.6$\pm$3.4  & 69.0$\pm$6.5 & 89.3$\pm$3.9  & 80.7$\pm$2.3  & \cellcolor[HTML]{EFEFEF}92.2$\pm$2.9          & 80.3$\pm$2.6  & 77.6$\pm$8.4 & 84.5$\pm$9.0  & 80.7$\pm$3.2  & \cellcolor[HTML]{EFEFEF}92.2$\pm$1.2          \\
Wood             & 99.6$\pm$0.1  & 98.8$\pm$0.3 & 99.3$\pm$0.1  & 100.0$\pm$0.0 & \cellcolor[HTML]{EFEFEF}99.6$\pm$0.3          & 99.7$\pm$0.1  & 99.0$\pm$0.1 & 99.5$\pm$0.2  & 100.0$\pm$0.0 & \cellcolor[HTML]{EFEFEF}99.7$\pm$0.1          & 99.7$\pm$0.2  & 99.1$\pm$0.0 & 99.5$\pm$0.2  & 99.9$\pm$0.1  & \cellcolor[HTML]{EFEFEF}99.5$\pm$0.1          \\
Zipper           & 96.9$\pm$0.5  & 95.5$\pm$0.9 & 97.2$\pm$0.3  & 96.8$\pm$1.8  & \cellcolor[HTML]{EFEFEF}99.0$\pm$0.6          & 98.2$\pm$0.8  & 95.4$\pm$1.0 & 97.8$\pm$1.0  & 98.3$\pm$0.4  & \cellcolor[HTML]{EFEFEF}99.3$\pm$0.5          & 98.6$\pm$0.4  & 96.2$\pm$0.8 & 99.1$\pm$0.7  & 98.5$\pm$0.2  & \cellcolor[HTML]{EFEFEF}98.5$\pm$0.3          \\ \midrule
Mean             & 90.6$\pm$0.8  & 88.1$\pm$1.7 & 92.2$\pm$1.5  & \underline{96.5$\pm$0.9}  & \cellcolor[HTML]{EFEFEF}\textbf{97.1$\pm$1.0} & 91.7$\pm$1.2  & 89.3$\pm$1.7 & 93.8$\pm$1.7  & \underline{97.0$\pm$0.7}  & \cellcolor[HTML]{EFEFEF}\textbf{97.9$\pm$0.7} & 92.5$\pm$1.2  & 90.5$\pm$1.6 & 94.5$\pm$1.5  & \underline{97.3$\pm$0.6}  & \cellcolor[HTML]{EFEFEF}\textbf{98.5$\pm$0.5} \\ \bottomrule
\end{tabular}
}
\caption{Comparison of image-level anomaly detection in terms of subset-wise AUPR on MVTec.}
\label{MVTec-I-PR}
\end{table*}

\begin{table*}[]
\centering
\resizebox{\linewidth}{!}{
\begin{tabular}{c|ccccc|ccccc|ccccc}
\toprule
MVTec & \multicolumn{5}{c|}{1-shot} & \multicolumn{5}{c|}{2-shot} & \multicolumn{5}{c}{4-shot} \\
Pixel-PRO  & SPADE    & PaDiM    & PatchCore & WinCLIP+  & \cellcolor[HTML]{EFEFEF}PromptAD          & SPADE    & PaDiM    & PatchCore & WinCLIP+  & \cellcolor[HTML]{EFEFEF}PromptAD          & SPADE    & PaDiM    & PatchCore & WinCLIP+  & \cellcolor[HTML]{EFEFEF}PromptAD          \\ \midrule
Bottle     & 91.1$\pm$0.4 & 89.8$\pm$0.8 & 93.5$\pm$0.3  & 91.2$\pm$0.4 & \cellcolor[HTML]{EFEFEF}93.6$\pm$0.1          & 91.8$\pm$0.5 & 91.7$\pm$0.2 & 93.9$\pm$0.3  & 91.8$\pm$0.3 & \cellcolor[HTML]{EFEFEF}93.9$\pm$0.2          & 92.5$\pm$0.1 & 92.2$\pm$0.2 & 94.0$\pm$0.2  & 91.6$\pm$0.2 & \cellcolor[HTML]{EFEFEF}94.5$\pm$0.2          \\
Cable      & 63.5$\pm$0.7 & 59.1$\pm$3.2 & 84.7$\pm$1.0  & 72.5$\pm$2.3 & \cellcolor[HTML]{EFEFEF}87.3$\pm$1.2          & 66.7$\pm$0.9 & 66.5$\pm$2.8 & 88.5$\pm$0.9  & 74.7$\pm$2.3 & \cellcolor[HTML]{EFEFEF}87.8$\pm$0.7          & 69.5$\pm$0.4 & 74.2$\pm$1.8 & 91.7$\pm$0.6  & 77.0$\pm$1.1 & \cellcolor[HTML]{EFEFEF}88.9$\pm$0.3          \\
Capsule    & 92.7$\pm$0.4 & 80.0$\pm$2.0 & 83.9$\pm$0.9  & 85.6$\pm$2.7 & \cellcolor[HTML]{EFEFEF}80.1$\pm$1.7          & 93.4$\pm$0.3 & 82.3$\pm$2.1 & 86.6$\pm$1.0  & 90.6$\pm$0.6 & \cellcolor[HTML]{EFEFEF}79.2$\pm$2.2          & 94.1$\pm$0.6 & 85.7$\pm$1.3 & 87.8$\pm$1.9  & 90.1$\pm$1.5 & \cellcolor[HTML]{EFEFEF}88.7$\pm$2.0          \\
Carpet     & 96.1$\pm$0.0 & 92.9$\pm$0.3 & 93.3$\pm$0.3  & 97.4$\pm$0.4 & \cellcolor[HTML]{EFEFEF}98.3$\pm$0.3          & 96.2$\pm$0.0 & 93.9$\pm$0.2 & 93.7$\pm$0.4  & 97.3$\pm$0.3 & \cellcolor[HTML]{EFEFEF}98.2$\pm$0.3          & 96.3$\pm$0.0 & 94.4$\pm$0.2 & 93.9$\pm$0.4  & 97.0$\pm$0.2 & \cellcolor[HTML]{EFEFEF}98.2$\pm$0.1          \\
Grid       & 67.7$\pm$1.9 & 41.2$\pm$4.6 & 21.7$\pm$9.5  & 90.5$\pm$2.7 & \cellcolor[HTML]{EFEFEF}94.3$\pm$1.0          & 72.1$\pm$1.5 & 45.1$\pm$3.6 & 23.7$\pm$3.8  & 92.8$\pm$2.5 & \cellcolor[HTML]{EFEFEF}95.0$\pm$0.8          & 78.0$\pm$1.5 & 55.5$\pm$3.4 & 30.4$\pm$4.6  & 93.6$\pm$0.6 & \cellcolor[HTML]{EFEFEF}93.8$\pm$0.7          \\
Hazelnut   & 94.9$\pm$0.3 & 85.7$\pm$1.9 & 88.3$\pm$1.3  & 93.7$\pm$0.9 & \cellcolor[HTML]{EFEFEF}92.9$\pm$0.5          & 95.6$\pm$0.2 & 89.4$\pm$0.9 & 89.8$\pm$1.3  & 94.2$\pm$0.3 & \cellcolor[HTML]{EFEFEF}93.4$\pm$0.5          & 95.6$\pm$0.1 & 90.4$\pm$0.7 & 92.0$\pm$0.3  & 94.2$\pm$0.3 & \cellcolor[HTML]{EFEFEF}95.2$\pm$0.5          \\
Leather    & 98.7$\pm$0.0 & 95.6$\pm$0.2 & 95.2$\pm$1.0  & 98.6$\pm$0.0 & \cellcolor[HTML]{EFEFEF}98.7$\pm$0.5          & 98.8$\pm$0.0 & 96.2$\pm$0.2 & 95.9$\pm$0.3  & 98.3$\pm$0.4 & \cellcolor[HTML]{EFEFEF}98.7$\pm$0.4          & 98.8$\pm$0.0 & 96.3$\pm$0.1 & 96.4$\pm$0.1  & 98.0$\pm$0.4 & \cellcolor[HTML]{EFEFEF}98.4$\pm$0.5          \\
Metal nut  & 73.4$\pm$1.1 & 38.1$\pm$1.6 & 66.7$\pm$2.9  & 84.7$\pm$1.1 & \cellcolor[HTML]{EFEFEF}83.1$\pm$0.6          & 78.1$\pm$1.8 & 48.2$\pm$5.0 & 79.6$\pm$4.2  & 86.7$\pm$0.8 & \cellcolor[HTML]{EFEFEF}87.7$\pm$1.1          & 81.2$\pm$1.4 & 54.0$\pm$8.8 & 83.8$\pm$5.5  & 89.4$\pm$0.1 & \cellcolor[HTML]{EFEFEF}87.6$\pm$0.3          \\
Pill       & 92.8$\pm$0.3 & 78.9$\pm$0.6 & 89.5$\pm$1.6  & 93.5$\pm$0.2 & \cellcolor[HTML]{EFEFEF}90.8$\pm$0.4          & 93.3$\pm$0.2 & 84.3$\pm$0.4 & 91.6$\pm$0.5  & 94.5$\pm$0.2 & \cellcolor[HTML]{EFEFEF}90.5$\pm$0.4          & 93.9$\pm$0.2 & 86.6$\pm$0.4 & 92.5$\pm$0.4  & 94.6$\pm$0.3 & \cellcolor[HTML]{EFEFEF}92.0$\pm$0.1          \\
Screw      & 85.0$\pm$0.8 & 51.6$\pm$1.7 & 68.1$\pm$1.3  & 82.3$\pm$1.1 & \cellcolor[HTML]{EFEFEF}78.1$\pm$1.7          & 87.2$\pm$1.2 & 69.5$\pm$2.1 & 69.0$\pm$2.1  & 84.1$\pm$0.5 & \cellcolor[HTML]{EFEFEF}74.7$\pm$1.4          & 89.5$\pm$1.3 & 72.3$\pm$0.8 & 72.4$\pm$3.1  & 86.3$\pm$1.8 & \cellcolor[HTML]{EFEFEF}86.7$\pm$2.5          \\
Tile       & 84.2$\pm$0.4 & 66.7$\pm$1.5 & 82.5$\pm$1.1  & 89.4$\pm$0.4 & \cellcolor[HTML]{EFEFEF}90.7$\pm$0.3          & 84.6$\pm$0.2 & 71.9$\pm$0.5 & 82.5$\pm$0.5  & 89.6$\pm$0.4 & \cellcolor[HTML]{EFEFEF}90.9$\pm$0.2          & 84.9$\pm$0.1 & 73.6$\pm$0.9 & 83.0$\pm$0.1  & 89.9$\pm$0.3 & \cellcolor[HTML]{EFEFEF}90.9$\pm$0.2          \\
Toothbrush & 83.5$\pm$1.3 & 82.1$\pm$1.5 & 79.0$\pm$2.4  & 85.3$\pm$1.0 & \cellcolor[HTML]{EFEFEF}90.1$\pm$0.5          & 87.4$\pm$1.1 & 83.3$\pm$2.6 & 81.0$\pm$0.7  & 84.7$\pm$1.4 & \cellcolor[HTML]{EFEFEF}91.6$\pm$0.5          & 89.0$\pm$1.1 & 87.1$\pm$1.7 & 85.5$\pm$3.0  & 86.0$\pm$3.3 & \cellcolor[HTML]{EFEFEF}91.3$\pm$0.2          \\
Transistor & 55.3$\pm$2.0 & 70.3$\pm$7.0 & 70.9$\pm$4.6  & 65.0$\pm$1.8 & \cellcolor[HTML]{EFEFEF}67.5$\pm$3.7          & 57.6$\pm$1.4 & 76.5$\pm$5.5 & 78.8$\pm$1.5  & 68.6$\pm$1.1 & \cellcolor[HTML]{EFEFEF}68.1$\pm$2.1          & 58.5$\pm$0.7 & 82.2$\pm$7.4 & 79.5$\pm$2.8  & 69.0$\pm$1.1 & \cellcolor[HTML]{EFEFEF}73.0$\pm$1.2          \\
Wood       & 92.9$\pm$0.1 & 86.5$\pm$0.6 & 87.1$\pm$1.0  & 91.0$\pm$0.6 & \cellcolor[HTML]{EFEFEF}92.4$\pm$0.8          & 93.1$\pm$0.1 & 88.0$\pm$0.2 & 86.8$\pm$1.4  & 91.8$\pm$0.6 & \cellcolor[HTML]{EFEFEF}91.6$\pm$0.6          & 93.2$\pm$0.1 & 88.4$\pm$0.2 & 87.7$\pm$0.4  & 91.7$\pm$0.3 & \cellcolor[HTML]{EFEFEF}91.4$\pm$0.5          \\
Zipper     & 86.8$\pm$0.6 & 81.7$\pm$2.0 & 91.2$\pm$1.1  & 86.0$\pm$1.7 & \cellcolor[HTML]{EFEFEF}81.0$\pm$1.4          & 89.0$\pm$0.4 & 85.6$\pm$0.7 & 92.8$\pm$0.4  & 86.4$\pm$1.6 & \cellcolor[HTML]{EFEFEF}86.4$\pm$0.5          & 90.1$\pm$0.2 & 87.2$\pm$0.8 & 93.4$\pm$0.2  & 86.9$\pm$0.7 & \cellcolor[HTML]{EFEFEF}87.5$\pm$0.6          \\ \midrule
Mean       & 83.9$\pm$0.7 & 73.3$\pm$2.0 & 79.7$\pm$2.0  & \underline{87.1$\pm$1.2} & \cellcolor[HTML]{EFEFEF}\textbf{87.9$\pm$1.0} & 85.7$\pm$0.7 & 78.2$\pm$1.8 & 82.3$\pm$1.3  & \underline{88.4$\pm$0.9} & \cellcolor[HTML]{EFEFEF}\textbf{88.5$\pm$0.8} & 87.0$\pm$0.5 & 81.3$\pm$1.9 & 84.3$\pm$1.6  & \underline{89.0$\pm$0.8} & \cellcolor[HTML]{EFEFEF}\textbf{90.5$\pm$0.7} \\ \bottomrule
\end{tabular}
}
\caption{Comparison of pixel-level anomaly detection in terms of subset-wise PRO on MVTec.}
\label{MVTec-P-PRO}
\end{table*}

\begin{table*}[]
\centering
\resizebox{\linewidth}{!}{
\begin{tabular}{c|ccccc|ccccc|ccccc}
\toprule
VisA & \multicolumn{5}{c|}{1-shot} & \multicolumn{5}{c|}{2-shot} & \multicolumn{5}{c}{4-shot} \\
Image-level AUPR & SPADE     & PaDiM     & PatchCore & WinCLIP+   & \cellcolor[HTML]{EFEFEF}PromptAD          & SPADE     & PaDiM     & PatchCore & WinCLIP+   & \cellcolor[HTML]{EFEFEF}PromptAD          & SPADE    & PaDiM    & PatchCore & WinCLIP+  & \cellcolor[HTML]{EFEFEF}PromptAD          \\ \midrule
Candle           & 86.5$\pm$4.3  & 69.2$\pm$3.9  & 86.6$\pm$2.3  & 93.6$\pm$1.5  & \cellcolor[HTML]{EFEFEF}93.7$\pm$2.9          & 90.7$\pm$3.2  & 72.8$\pm$1.0  & 86.8$\pm$1.7  & 95.1$\pm$1.1  & \cellcolor[HTML]{EFEFEF}93.6$\pm$1.1          & 92.6$\pm$1.9 & 72.5$\pm$1.1 & 88.9$\pm$1.1  & 95.3$\pm$0.4 & \cellcolor[HTML]{EFEFEF}92.9$\pm$1.1          \\
Capsules         & 79.4$\pm$4.9  & 63.4$\pm$5.7  & 72.3$\pm$5.3  & 89.9$\pm$2.5  & \cellcolor[HTML]{EFEFEF}90.1$\pm$1.9          & 79.9$\pm$5.8  & 63.4$\pm$2.0  & 73.6$\pm$4.7  & 88.9$\pm$0.7  & \cellcolor[HTML]{EFEFEF}88.3$\pm$2.6          & 81.1$\pm$4.5 & 63.0$\pm$2.3 & 78.4$\pm$3.1  & 91.5$\pm$1.4 & \cellcolor[HTML]{EFEFEF}89.8$\pm$1.0          \\
Cashew           & 97.9$\pm$0.4  & 78.2$\pm$5.7  & 94.6$\pm$2.0  & 97.2$\pm$0.2  & \cellcolor[HTML]{EFEFEF}97.6$\pm$0.7          & 98.6$\pm$0.6  & 86.1$\pm$2.2  & 96.9$\pm$0.3  & 97.3$\pm$0.2  & \cellcolor[HTML]{EFEFEF}97.4$\pm$0.5          & 98.3$\pm$0.6 & 88.4$\pm$2.0 & 96.5$\pm$0.7  & 97.7$\pm$0.4 & \cellcolor[HTML]{EFEFEF}97.0$\pm$1.1          \\
Chewinggum       & 96.4$\pm$0.9  & 79.8$\pm$3.6  & 98.9$\pm$0.1  & 99.0$\pm$0.3  & \cellcolor[HTML]{EFEFEF}99.1$\pm$0.7          & 97.1$\pm$0.4  & 89.5$\pm$1.9  & 99.1$\pm$0.2  & 98.9$\pm$0.3  & \cellcolor[HTML]{EFEFEF}98.4$\pm$0.2          & 97.1$\pm$0.6 & 88.5$\pm$3.2 & 99.3$\pm$0.1  & 99.0$\pm$0.1 & \cellcolor[HTML]{EFEFEF}98.5$\pm$0.3          \\
Fryum            & 89.8$\pm$1.8  & 74.5$\pm$2.9  & 87.6$\pm$2.4  & 94.7$\pm$1.0  & \cellcolor[HTML]{EFEFEF}93.8$\pm$1.0          & 94.5$\pm$2.3  & 81.0$\pm$5.4  & 92.1$\pm$1.3  & 95.8$\pm$0.2  & \cellcolor[HTML]{EFEFEF}96.0$\pm$0.7          & 95.8$\pm$1.0 & 81.5$\pm$3.0 & 95.0$\pm$0.6  & 96.0$\pm$0.3 & \cellcolor[HTML]{EFEFEF}93.6$\pm$0.5          \\
Macaroni1        & 61.9$\pm$11.2 & 60.4$\pm$2.9  & 67.8$\pm$3.4  & 84.9$\pm$1.2  & \cellcolor[HTML]{EFEFEF}86.3$\pm$1.9          & 64.5$\pm$9.5  & 63.1$\pm$4.3  & 74.9$\pm$5.2  & 84.7$\pm$1.5  & \cellcolor[HTML]{EFEFEF}91.1$\pm$1.7          & 60.2$\pm$2.7 & 64.9$\pm$2.1 & 82.1$\pm$3.5  & 86.5$\pm$0.6 & \cellcolor[HTML]{EFEFEF}89.2$\pm$1.3          \\
Macaroni2        & 52.7$\pm$4.2  & 51.7$\pm$5.0  & 54.9$\pm$3.2  & 68.4$\pm$1.8  & \cellcolor[HTML]{EFEFEF}72.5$\pm$2.5          & 55.9$\pm$3.1  & 52.7$\pm$1.5  & 57.2$\pm$2.6  & 70.4$\pm$1.8  & \cellcolor[HTML]{EFEFEF}84.7$\pm$1.6          & 51.9$\pm$2.3 & 54.9$\pm$2.5 & 60.2$\pm$3.0  & 69.6$\pm$2.8 & \cellcolor[HTML]{EFEFEF}82.2$\pm$1.0          \\
PCB1             & 84.9$\pm$3.7  & 68.6$\pm$2.4  & 72.1$\pm$2.5  & 76.5$\pm$19.0 & \cellcolor[HTML]{EFEFEF}88.0$\pm$11.3         & 83.8$\pm$2.1  & 60.4$\pm$7.7  & 72.6$\pm$16.4 & 78.3$\pm$4.3  & \cellcolor[HTML]{EFEFEF}80.9$\pm$6.3          & 83.2$\pm$7.2 & 77.4$\pm$2.9 & 81.0$\pm$9.2  & 87.7$\pm$1.7 & \cellcolor[HTML]{EFEFEF}90.1$\pm$3.6          \\
PCB2             & 74.9$\pm$2.9  & 63.3$\pm$1.2  & 84.4$\pm$0.4  & 64.9$\pm$3.3  & \cellcolor[HTML]{EFEFEF}75.4$\pm$2.7          & 71.7$\pm$6.6  & 68.9$\pm$2.6  & 86.6$\pm$1.1  & 65.8$\pm$4.0  & \cellcolor[HTML]{EFEFEF}73.0$\pm$4.8          & 74.2$\pm$5.0 & 75.0$\pm$1.7 & 86.2$\pm$1.0  & 71.3$\pm$3.4 & \cellcolor[HTML]{EFEFEF}75.3$\pm$2.5          \\
PCB3             & 75.5$\pm$2.1  & 52.3$\pm$10.8 & 84.6$\pm$1.5  & 73.5$\pm$1.6  & \cellcolor[HTML]{EFEFEF}75.2$\pm$3.8          & 78.3$\pm$5.2  & 65.2$\pm$3.8  & 86.1$\pm$0.5  & 80.9$\pm$1.6  & \cellcolor[HTML]{EFEFEF}82.8$\pm$2.2          & 81.0$\pm$3.6 & 64.5$\pm$2.4 & 88.3$\pm$1.1  & 84.8$\pm$1.8 & \cellcolor[HTML]{EFEFEF}83.5$\pm$1.6          \\
PCB4             & 92.9$\pm$1.6  & 74.7$\pm$2.6  & 92.8$\pm$3.1  & 78.5$\pm$15.5 & \cellcolor[HTML]{EFEFEF}90.5$\pm$1.2          & 81.9$\pm$11.2 & 67.6$\pm$11.9 & 93.2$\pm$3.4  & 72.5$\pm$16.2 & \cellcolor[HTML]{EFEFEF}94.5$\pm$2.9          & 94.8$\pm$2.9 & 84.0$\pm$2.0 & 94.9$\pm$1.2  & 85.6$\pm$8.9 & \cellcolor[HTML]{EFEFEF}97.5$\pm$1.3          \\
Pipe fryum       & 88.3$\pm$2.0  & 79.2$\pm$1.5  & 95.4$\pm$0.6  & 98.6$\pm$0.5  & \cellcolor[HTML]{EFEFEF}98.3$\pm$0.1          & 88.1$\pm$1.7  & 84.5$\pm$1.7  & 96.8$\pm$0.7  & 99.0$\pm$0.3  & \cellcolor[HTML]{EFEFEF}99.1$\pm$0.1          & 88.8$\pm$1.0 & 89.8$\pm$1.7 & 98.3$\pm$0.3  & 99.2$\pm$0.2 & \cellcolor[HTML]{EFEFEF}99.3$\pm$0.1          \\ \midrule
Mean             & 82.0$\pm$3.3  & 68.3$\pm$4.0  & 82.8$\pm$2.3  & \underline{85.1$\pm$4.0}  & \cellcolor[HTML]{EFEFEF}\textbf{88.4$\pm$2.6} & 82.3$\pm$4.3  & 71.6$\pm$3.8  & 84.8$\pm$3.2  & \underline{85.8$\pm$2.7}  & \cellcolor[HTML]{EFEFEF}\textbf{90.0$\pm$2.1} & 83.4$\pm$2.7 & 75.6$\pm$2.2 & 87.5$\pm$2.1  & \underline{88.8$\pm$1.8} & \cellcolor[HTML]{EFEFEF}\textbf{90.8$\pm$1.3} \\ \bottomrule
\end{tabular}
}
\caption{Comparison of image-level anomaly detection in terms of subset-wise AUPR on VisA.}
\label{VisA-I-PR}
\end{table*}

\begin{table*}[]
\centering
\resizebox{\linewidth}{!}{
\begin{tabular}{c|ccccc|ccccc|ccccc}
\toprule
VisA & \multicolumn{5}{c|}{1-shot} & \multicolumn{5}{c|}{2-shot} & \multicolumn{5}{c}{4-shot} \\
Pixel-PRO  & SPADE    & PaDiM    & PatchCore & WinCLIP+           & \cellcolor[HTML]{EFEFEF}PromptAD          & SPADE    & PaDiM    & PatchCore & WinCLIP+           & \cellcolor[HTML]{EFEFEF}PromptAD & SPADE    & PaDiM    & PatchCore & WinCLIP+           & \cellcolor[HTML]{EFEFEF}PromptAD \\ \midrule
Candle     & 95.6$\pm$0.5 & 81.5$\pm$5.3 & 92.6$\pm$0.4  & 94.0$\pm$0.4          & \cellcolor[HTML]{EFEFEF}91.8$\pm$1.2          & 95.6$\pm$0.4 & 87.3$\pm$1.2 & 93.4$\pm$0.6  & 94.2$\pm$0.2          & \cellcolor[HTML]{EFEFEF}91.6$\pm$0.7 & 95.7$\pm$0.1 & 88.3$\pm$0.7 & 94.1$\pm$0.4  & 94.4$\pm$0.2          & \cellcolor[HTML]{EFEFEF}90.6$\pm$0.5 \\
Capsules   & 83.1$\pm$1.1 & 30.6$\pm$1.1 & 66.6$\pm$4.5  & 73.6$\pm$3.5          & \cellcolor[HTML]{EFEFEF}70.0$\pm$1.6          & 85.4$\pm$3.1 & 38.4$\pm$3.7 & 67.9$\pm$2.3  & 75.9$\pm$1.9          & \cellcolor[HTML]{EFEFEF}70.8$\pm$2.5 & 89.0$\pm$1.2 & 43.3$\pm$2.0 & 69.0$\pm$3.2  & 77.0$\pm$1.4          & \cellcolor[HTML]{EFEFEF}72.4$\pm$3.2 \\
Cashew     & 89.8$\pm$1.1 & 73.4$\pm$2.1 & 90.8$\pm$0.2  & 91.1$\pm$0.8          & \cellcolor[HTML]{EFEFEF}92.3$\pm$0.5          & 90.4$\pm$0.5 & 78.4$\pm$2.7 & 91.4$\pm$1.0  & 90.4$\pm$0.6          & \cellcolor[HTML]{EFEFEF}92.7$\pm$1.8 & 90.4$\pm$0.6 & 81.2$\pm$2.8 & 92.1$\pm$0.3  & 91.3$\pm$0.9          & \cellcolor[HTML]{EFEFEF}92.8$\pm$2.0 \\
Chewinggum & 73.9$\pm$1.2 & 58.1$\pm$0.6 & 78.2$\pm$1.3  & 91.0$\pm$0.5          & \cellcolor[HTML]{EFEFEF}89.8$\pm$1.3          & 73.8$\pm$1.1 & 63.7$\pm$2.4 & 78.0$\pm$0.4  & 90.9$\pm$0.7          & \cellcolor[HTML]{EFEFEF}87.8$\pm$1.3 & 72.7$\pm$0.9 & 67.2$\pm$1.8 & 79.3$\pm$0.8  & 91.0$\pm$0.4          & \cellcolor[HTML]{EFEFEF}89.4$\pm$0.6 \\
Fryum      & 83.7$\pm$1.2 & 71.1$\pm$1.6 & 78.7$\pm$2.3  & 89.1$\pm$1.0          & \cellcolor[HTML]{EFEFEF}83.5$\pm$3.1          & 84.5$\pm$0.9 & 71.2$\pm$0.8 & 81.4$\pm$2.8  & 89.3$\pm$0.2          & \cellcolor[HTML]{EFEFEF}86.2$\pm$3.1 & 86.2$\pm$0.9 & 73.2$\pm$1.3 & 81.0$\pm$1.2  & 89.7$\pm$0.5          & \cellcolor[HTML]{EFEFEF}80.3$\pm$0.7 \\
Macaroni1  & 92.0$\pm$0.6 & 62.2$\pm$4.4 & 83.4$\pm$1.3  & 84.6$\pm$2.3          & \cellcolor[HTML]{EFEFEF}87.5$\pm$1.9          & 93.9$\pm$0.8 & 71.8$\pm$2.4 & 86.2$\pm$4.6  & 85.2$\pm$1.4          & \cellcolor[HTML]{EFEFEF}90.6$\pm$1.5 & 95.1$\pm$0.4 & 76.6$\pm$2.1 & 89.6$\pm$0.7  & 86.8$\pm$0.8          & \cellcolor[HTML]{EFEFEF}91.5$\pm$1.3 \\
Macaroni2  & 80.0$\pm$3.3 & 54.9$\pm$3.6 & 66.0$\pm$3.0  & 89.3$\pm$2.4          & \cellcolor[HTML]{EFEFEF}80.6$\pm$1.5          & 81.7$\pm$1.5 & 65.6$\pm$3.4 & 67.2$\pm$6.5  & 88.6$\pm$1.7          & \cellcolor[HTML]{EFEFEF}82.7$\pm$1.0 & 86.0$\pm$0.8 & 65.9$\pm$1.5 & 78.3$\pm$0.9  & 90.5$\pm$1.3          & \cellcolor[HTML]{EFEFEF}87.2$\pm$0.6 \\
PCB1       & 81.3$\pm$5.7 & 63.9$\pm$1.8 & 79.0$\pm$10.7 & 82.5$\pm$6.0          & \cellcolor[HTML]{EFEFEF}89.2$\pm$13.1         & 87.2$\pm$2.3 & 68.4$\pm$4.1 & 86.1$\pm$1.7  & 83.8$\pm$5.0          & \cellcolor[HTML]{EFEFEF}90.2$\pm$7.5 & 88.0$\pm$2.7 & 70.2$\pm$3.3 & 88.1$\pm$2.6  & 87.9$\pm$2.1          & \cellcolor[HTML]{EFEFEF}90.2$\pm$6.0 \\
PCB2       & 83.7$\pm$0.6 & 64.4$\pm$3.8 & 80.9$\pm$0.5  & 73.6$\pm$1.5          & \cellcolor[HTML]{EFEFEF}79.3$\pm$2.0          & 85.5$\pm$1.0 & 72.9$\pm$3.4 & 82.9$\pm$1.8  & 76.2$\pm$0.9          & \cellcolor[HTML]{EFEFEF}79.3$\pm$1.9 & 87.0$\pm$0.5 & 71.9$\pm$2.6 & 83.7$\pm$1.0  & 78.0$\pm$1.3          & \cellcolor[HTML]{EFEFEF}76.3$\pm$1.6 \\
PCB3       & 84.3$\pm$1.0 & 69.0$\pm$1.2 & 78.1$\pm$2.0  & 79.5$\pm$2.5          & \cellcolor[HTML]{EFEFEF}84.0$\pm$1.4          & 86.1$\pm$0.6 & 74.0$\pm$2.3 & 82.2$\pm$1.1  & 82.3$\pm$1.8          & \cellcolor[HTML]{EFEFEF}84.7$\pm$1.1 & 87.7$\pm$0.6 & 77.2$\pm$0.8 & 84.4$\pm$1.9  & 84.2$\pm$1.0          & \cellcolor[HTML]{EFEFEF}85.0$\pm$1.3 \\
PCB4       & 66.9$\pm$2.0 & 59.1$\pm$1.8 & 77.9$\pm$3.1  & 76.6$\pm$4.1          & \cellcolor[HTML]{EFEFEF}78.8$\pm$2.3          & 69.3$\pm$1.1 & 62.6$\pm$3.6 & 79.5$\pm$4.8  & 81.7$\pm$1.2          & \cellcolor[HTML]{EFEFEF}78.3$\pm$2.6 & 74.7$\pm$1.0 & 67.9$\pm$2.6 & 83.5$\pm$2.5  & 84.2$\pm$0.7          & \cellcolor[HTML]{EFEFEF}83.4$\pm$2.4 \\
Pipe fryum & 94.3$\pm$0.5 & 83.9$\pm$0.8 & 93.6$\pm$0.5  & 96.1$\pm$0.6          & \cellcolor[HTML]{EFEFEF}95.2$\pm$0.4          & 95.0$\pm$0.2 & 86.9$\pm$0.9 & 94.5$\pm$0.4  & 96.2$\pm$0.6          & \cellcolor[HTML]{EFEFEF}94.8$\pm$0.5 & 95.0$\pm$0.3 & 88.7$\pm$1.3 & 95.0$\pm$0.5  & 96.6$\pm$0.2          & \cellcolor[HTML]{EFEFEF}95.3$\pm$0.3 \\ \midrule
Mean       & \underline{84.1$\pm$1.6} & 64.3$\pm$2.4 & 80.5$\pm$2.5  & \textbf{85.1$\pm$2.1} & \cellcolor[HTML]{EFEFEF}\textbf{85.1$\pm$2.5} & 85.7$\pm$1.1 & 70.1$\pm$2.6 & 82.6$\pm$2.3  & \textbf{86.2$\pm$1.4} & \cellcolor[HTML]{EFEFEF}\underline{85.8$\pm$2.1} & \underline{87.3$\pm$0.8} & 72.6$\pm$1.9 & 84.9$\pm$1.4  & \textbf{87.6$\pm$0.9} & \cellcolor[HTML]{EFEFEF}86.2$\pm$1.7 \\ \bottomrule
\end{tabular}
}
\caption{Comparison of pixel-level anomaly detection in terms of subset-wise PRO on VisA.}
\label{VisA-P-PRO}
\end{table*}

\end{document}